\documentclass[journal,twoside,web]{ieeecolor}
\usepackage{tmi}
\usepackage[caption=false]{subfig}
\usepackage[nospace]{cite}
\usepackage{amsmath,amssymb,amsfonts}
\usepackage{algorithmic}
\usepackage{graphicx}
\usepackage{textcomp}
\usepackage[super]{nth}

\newcommand{\etal}{\emph{et al.}}
\usepackage[mathlines,switch]{lineno}
\usepackage[bookmarks=false]{hyperref}

\DeclareMathOperator*{\argmax}{arg\,max}
\DeclareMathOperator*{\argmin}{arg\,min}
\def\BibTeX{{\rm B\kern-.05em{\sc i\kern-.025em b}\kern-.08em T\kern-.1667em\lower.7ex\hbox{E}\kern-.125emX}}
\makeatletter
\def\fnum@figure{\textcolor{subsectioncolor}{\sf Fig.~\thefigure}}
\def\fnum@table{\textcolor{subsectioncolor}{\sf TABLE~\thetable}}
\makeatother
\begin{document}
\bstctlcite{IEEEexample:BSTcontrol}
\title{Adapt Everywhere: Unsupervised Adaptation of Point-Clouds and Entropy Minimisation for Multi-modal Cardiac Image Segmentation}

\author{Sulaiman Vesal,
        Mingxuan Gu,
        Ronak Kosti,
        Andreas Maier~\IEEEmembership{Member,~IEEE}, and
        Nishant Ravikumar
        
\thanks{S. Vesal, M. Gu, R.Kosti, and A. Maier are with the Pattern Recognition Lab, Friedrich-Alexander-University Erlangen-Nuremberg, Germany. (E-mail: sulaiman.vesal@fau.de)}
\thanks{N. Ravikumar is with CISTIB, Centre for Computational Imaging and Simulation Technologies in Biomedicine, School of Computing, LICAMM Leeds Institute of Cardiovascular and Metabolic Medicine, School of Medicine, University of Leeds, United Kingdom.}
\thanks{The work described in this paper was partially supported by the project EFI-BIG-THERA: Integrative ‘BigData Modeling’ for the development of novel therapeutic approaches for breast cancer. The authors would also like to thank NVIDIA for donating a Titan X-Pascal GPU.}
}

\maketitle
\begin{abstract}
Deep learning models are sensitive to domain shift phenomena. A model trained on images from one domain cannot generalise well when tested on images from a different domain, despite capturing similar anatomical structures. It is mainly because the data distribution between the two domains is different. Moreover, creating annotation for every new modality is a tedious and time-consuming task, which also suffers from high inter- and intra- observer variability. Unsupervised domain adaptation (UDA) methods intend to reduce the gap between source and target domains by leveraging source domain labelled data to generate labels for the target domain. However, current state-of-the-art (SOTA) UDA methods demonstrate degraded performance when there is insufficient data in source and target domains. In this paper, we present a novel UDA method for multi-modal cardiac image segmentation. The proposed method is based on adversarial learning and adapts network features between source and target domain in different spaces. The paper introduces an end-to-end framework that integrates: a) entropy minimisation, b) output feature space alignment and c) a novel point-cloud shape adaptation based on the latent features learned by the segmentation model. We validated our method on two cardiac datasets by adapting from the annotated source domain, bSSFP-MRI (balanced Steady-State Free Procession-MRI), to the unannotated target domain, LGE-MRI (Late-gadolinium enhance-MRI), for the multi-sequence dataset; and from MRI (source) to CT (target) for the cross-modality dataset. The results highlighted that by enforcing adversarial learning in different parts of the network, the proposed method delivered promising performance, compared to other SOTA methods.
\end{abstract}


\section{Introduction}
\label{sec:introduction}
\IEEEPARstart{M}{yocardial} infarction (MI) is a cardiovascular disease with a high percentage of mortality and morbidity rate worldwide~\cite{KIM20091, 10.1093/eurheartj/ehx628}.  For the diagnosis and treatment of patients with MI, there is a need for modelling of the ventricle blood pools and accurate analysis of myocardium using different imaging modalities~\cite{8458220}. Cardiac magnetic resonance (CMR) imaging modalities are regularly used in the clinical diagnosis to provide anatomical morphology and operational information of the heart. A comprehensive diagnosis involves different types of CMR sequences which provide complementary information to each other. Among them LGE images are desirable to determine the presence, location, and extent of MI ~\cite{10.1093/ehjci/jev123}. However, manual contouring is usually time-consuming, tiresome, and subjected to inter- and intra-observer variations~\cite{zhuang2}. Due to the generation of multiple imaging modalities, there is a substantial clinical need to develop a multi-modal CMR segmentation system that can generalise well across different modalities~\cite{yan2019domain}.

Recently, Convolutional Neural Network (CNN) based methods have been widely used for medical image analysis in detection, segmentation~\cite{chen2019synergistic,liu2019cross}, and tracking of anatomical structures. Such methods are often generic and can be extended from one imaging modality to another by fine-tuning or re-training on the target imaging modality. However, to achieve satisfactory performance, a sufficient number of annotated target training images are required. In practice, it is often difficult to accumulate enough training images for a new imaging modality not well established in clinical practice yet.
Synthesising or data augmentation is often used to support training data in the hope that they can boost the generalisation capability of a trained deep learning model. However, the distribution gap between synthesised data and real data often determines the success of such an approach. 
 
Taking advantage of unlabelled data from other modalities is quite challenging due to the high data distribution discrepancy. Recent advances have used generative adversarial networks (GAN)~\cite{DCGAN} to formulate it as an image-to-image translation task. These methods require pixel-to-pixel correspondence between two domains to build a direct cross-modality segmentation model. However, multi-modal medical images are mostly in 3D and do not have cross-modal paired data. A method to learn from unpaired data is clinically more desirable.  Anatomical shape in medical images and volumes contain diagnostic information, so it is important to maintain translation invariance. However, GAN models that are trained without paired data do not guarantee this requirement due to the lack of direct reconstruction and the reliance on discriminators. Therefore, adapting features between different domains can avoid any complex mapping process between paired images \cite{Chen2019UnsupervisedMS, CYCADA}.

In this work, we propose a new scheme to leverage entropy and shape information available in the source and target domain for UDA. We hypothesise that introducing additional shape information using point-clouds along with entropy adaptation brings complementary effects to further bridge the performance gap between source and target at test time. To this end, we transform the segmentation network such that the shape information in the form of point-clouds is embedded into a dedicated deep architecture using an auxiliary point-cloud regression task. Point-clouds for cardiac shape estimation, operating as an additional source domain supervision in our framework (only available while training), will be considered as furnished information. Another challenge is to incorporate 2D point-cloud regression into UDA learning efficiently. We achieve this by introducing a new point-cloud adversarial training protocol based on PointNet discriminator. The proposed approach permits accurate segmentation of cardiac images without any annotation.

\noindent In summary, our main contributions are threefold: 
\begin{itemize}
    \item \textit{First}, we propose a novel UDA method based on adversarial learning that adapts the features between source and target domains in different spaces. Our network incorporates entropy minimisation, output space alignment, and point-cloud adaptation to tackle drastic domain shift. To the best of our knowledge, it is the first study that employed shape-prior information using point-clouds for UDA. 
    \item \textit{Second}, we present a novel multi-task model for point-set generation from the latent representation of the segmentation network. It allows multi-task learning with point-set network acting as a surface shape learning mechanism and consequently improves the segmentation network for the new domain.
    \item \textit{Third}, we validate the proposed point-cloud UDA  on two cardiac image segmentation tasks, including multi-sequences CMR (bSSFP, T2-weighted, and LGE) and cross-modal cardiac CT and MRI. The proposed method achieved promising results. Code and models available at: \url{https://github.com/sulaimanvesal/PointCloudUDA}
\end{itemize}

While working with multi-sequence data, MS-CMRSeg~\cite{zhuang2020cardiac}, the source and target domains are bSSFP-MRI (we combine bSSFP and T2 sequences into source domain) and LGE-MRI; whereas for cross-modal data, MM-WHS~\cite{MMWHS}, source and target domains are MRI and CT, respectively.

The rest of the paper is organised as follows: Section II reviews related work; Section III describes our proposed entropy and shape-aware UDA method; Section IV deals with comprehensive evaluation and analysis of our experiments; Section V discusses the statistical analysis and limitation of the proposed method, and Section VI concludes our work.

\section{Related Work}
Recently, UDA methods have been explored in the field of medical imaging to deal with the performance degradation caused by the domain shift. Existing methods on UDA generally suggest aligning the source and target domain distributions from three perspectives. First category is the image-level alignment, which transforms the image appearance between domains with an image-to-image transformation model~\cite{Zhu2017UnpairedCYCLEGAN, Russo_2018_CVPR, zhang2018task, Zhao2019, Chen2019UnsupervisedMS, CEFA2019}. The second category focuses on feature alignment, aiming to extract domain-invariant features usually by minimising feature distance between domains via adversarial learning~\cite{TsaiCVPR2018, vu2019advent, wang2019patch}. And, a recent third category focuses on alignment of feature-level and image-level information~\cite{chen2019synergistic}.

\noindent \textbf{Image-level domain adaptation:} Image-based UDA methods are developed based on unpaired image-to-image translation algorithms, which mainly use CycleGAN~\cite{Zhu2017UnpairedCYCLEGAN} and MUNIT~\cite{huang2018multimodal}. Bousmalis \emph{et al.}~\cite{shanis2019intramodality} aligned the image appearance between two modalities using the pixel-to-pixel transformation, employing domain adaptation at input space. In such models, domain adaptation relies highly on the quality of stylised images, which are often not perfect. Zhang \emph{et al.}~\cite{zhang2018translating} used cycle and shape-consistency adversarial networks for multi-modal brain MRI segmentation. In~\cite{liu2019cross}, the authors again used image translation to synthesise the data and incorporate it with an attention-based neural network. Delisle \emph{et al.}~\cite{delisle2019adversarial} developed an adversarial method to tackle UDA segmentation from a normalisation perspective.

\noindent \textbf{Feature-level domain adaptation:} 
In the feature-level, Kamnitsas \emph{et al.}~\cite{kamnitsas2017unsupervised} proposed a UDA for brain lesion segmentation, which learned domain-invariant features with a discriminator that predicts the input image domain. In cross-modal segmentation with drastic differences between the source and target domain, Q. Dou \emph{et al.}~\cite{dou2019pnp} fine-tuned specific feature layers, and employed adversarial loss for supervised feature learning. Recently, output space alignment is also exploited to incorporate the spatial and structural geometry information of predictions~\cite{TsaiCVPR2018}. Wang \emph{et al.}~\cite{BEAL2019} proposed a method to adversarially adapt entropy and boundary of Fundus images of different vendors. The new depth-aware adaptation scheme, DADA learning~\cite{Vu_2019_ICCV}, simultaneously aligns segmentation and depth-based information of source and target while being aware of scene geometry.

\noindent \textbf{Image and Feature domain adaptation:} CyCADA~\cite{CYCADA} poses UDA as style transfer with adversarial learning to bridge the gap in appearance between the source and target domains, while simultaneously aligning the image and latent feature spaces independently. To avoid divergence of semantics, they enforce cycle consistency during adaptation of the corresponding domains. Chen \emph{et al.}~\cite{chen2019synergistic} proposed a domain adaptation framework, SIFA, which considers both feature and image-level adaptation, concurrently for MRI to CT segmentation. Their network is built upon a CycleGAN, with additional discriminators to emphasise on feature-level adaptation, and image domain separation. They extended SIFA to Bidirectional-SIFA~\cite{chen2020unsupervised} by adding deeply supervised feature alignment and exploring adaptation in a bi-directional fashion ($MRI\Leftrightarrow CT$) achieving SOTA on multi-modal medical image segmentation.

\noindent \textbf{Multi-sequence domain adaptation:} For multi-sequence cardiac MRI segmentation, Chen \emph{et al.}~\cite{Chen2019UnsupervisedMS} proposed a network (bSSFP $\rightarrow$ LGE) based on multi-modal unsupervised image-to-image translation (MUNIT) network~\cite{huang2018multimodal}. The method transferred style, shape, and appearance from bSSFP images to LGE to generate synthetic samples, to train the segmentation network.  Wang \emph{et al.}~\cite{Wang2019} proposed a fully end-to-end unsupervised method based on adversarial training to minimise discrepancies in both the feature and output space. Roth \emph{et al.}~\cite{Roth2019} couples classical methods of multi-atlas label fusion with deep learning by formulating noisy labels for unlabelled LGE images using the registration technique.

Most of the previous methods fail to produce reliable predictions when the target images are noisy, visually different, or without clear boundaries. Cai \emph{et al.}~\cite{cai2019end} proposed using point-clouds to incorporate shape information during training, where their segmentation model is aware of the shape and topology of organs. Their shape learning multi-task network uses multi-scale features from the segmentation model to generate organ surface point-clouds with more details. They also incorporated a discriminator to improve the generation of point-clouds with fewer outliers in an adversarial fashion. As a result, the shape learning step acts as an auxiliary task to provide complementary information for the segmentation model and improves organ segmentation. Clearly, the shape information for cardiac ventricle segmentation is also an essential factor. Therefore, developing an effective UDA method to improve the prediction performance on the shape of the overall ventricles of the target domain images remains a challenge.

\section{Entropy and Shape-Aware Domain Adaptation}
\textbf{Problem Definition:} In UDA for semantic segmentation, we are given a set of source images, bSSFP-MRI (multi-sequence data) or MRI (multi-modal data), and their corresponding mask labels in the source domain $\mathbb{D}_{s} = {(\mathbf{I}_{i}^{s}, \mathbf{Y}_{i}^{s})}_{i=1}^{m_{s}}$, where $\mathbf{I}_{i}^{s} \in \mathcal{R} ^{w \times h \times 3}$, $\mathbf{Y}_{i}^{s} \in \mathcal{R} ^{w \times h \times c}$ and $m_{s}$ is the number of source images. In the target domain, LGE-MRI (multi-sequence data) or CT (multi-modal data), we are given unlabelled target images $\mathbb{D}_{t} = {(\mathbf{I}_{i}^{t})}_{i=1}^{m_{t}}$, where $m_{t}$ is the number of target images. Our goal is to train a supervised model on $\mathbb{D}_{s}$ and incorporate information from $\mathbb{D}_{t}$ to reduce the gap between two domains, and improve segmentation accuracy on $\mathbb{D}_{t}$ in an unsupervised manner. Due to domain shift, images across domains usually present different data distribution; the goal is to bring distributions of both the domains closer.

\subsection{Overview of the Proposed Method} 
We hypothesise that offering additional shape information using point-clouds along with feature adaptation brings complementary effects to bridge the performance gap between source and target at test time. Starting from an existing segmentation network, we insert additional modules: \emph{(1)} to predict object point-clouds as supplementary output, and \emph{(2)} to feed the information exploited by this auxiliary task back to the main network. To align the features for $\mathbb{D}_{s}$ and  $\mathbb{D}_{t}$, the most common way is applying adversarial learning directly in feature space, such that a discriminator learns to differentiate the domain space of the features. However, due to the high dimensionality of the feature space, it is difficult to align the features directly. Instead, we enhance the domain-invariance of feature distributions by: \textit{(1)} using adversarial learning via entropy minimisation, \textit{(2)} aligning objects' shape information with point-cloud data, and \textit{(3)} discriminating the structured output space. 

\begin{figure*}[ht]
    \centering
    \includegraphics[width=0.90\textwidth]{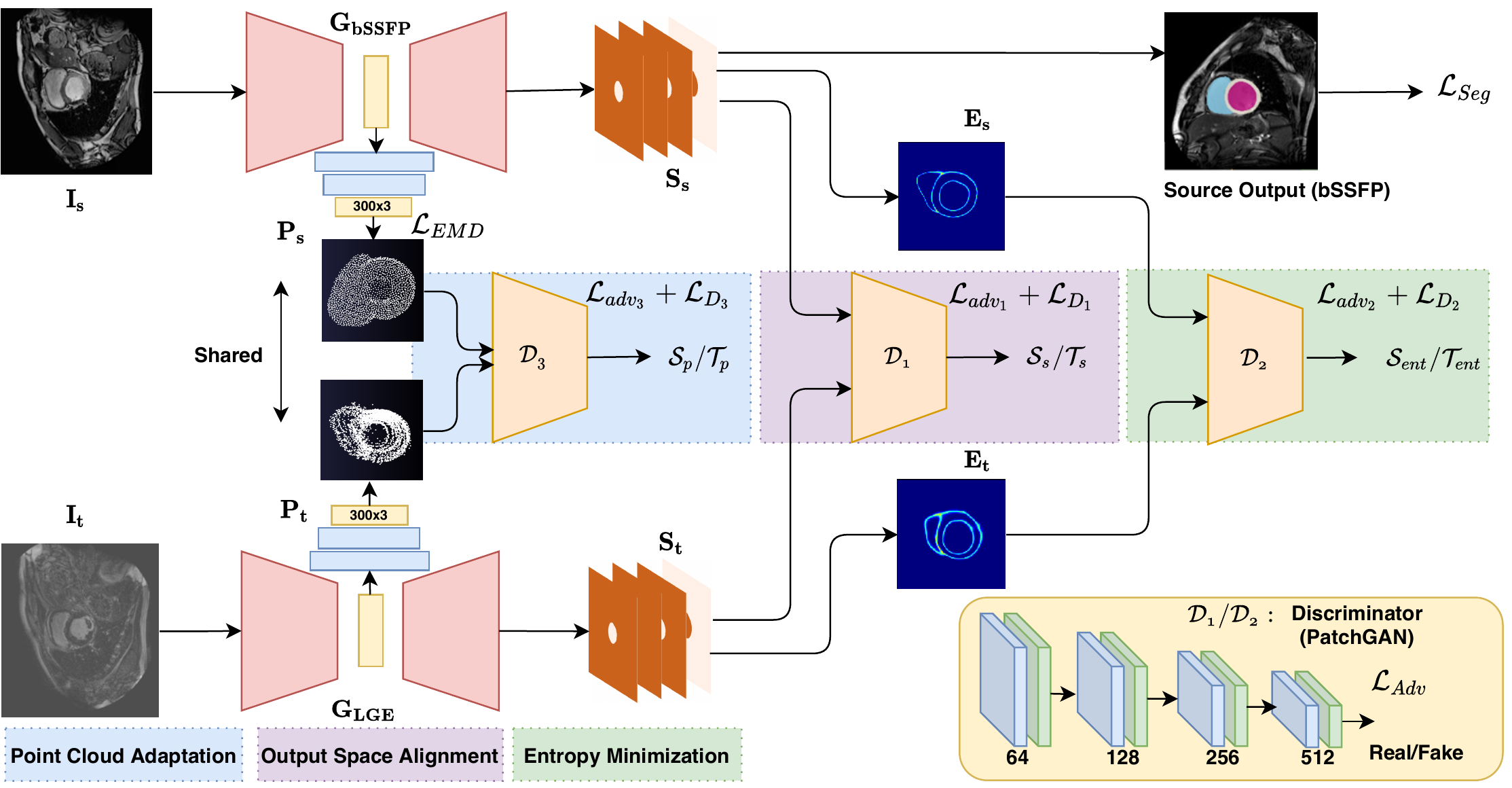}
    \caption{Overview of the proposed UDA segmentation framework. A single image from the source domain or the target domain fed into its corresponding domain-specific DR-UNet segmentation network (depicted in Fig. \ref{figSegNet}), which has shared weights. The DR-UNet encoder extracts high-level features for both the source and target domains. Then the features are sent to the decoder for segmentation and point-net to generate point-cloud. The point-cloud is fed to $\mathbf{D_{3}}$ for shape alignment (depicted in Fig. \ref{figpointdisc}). The output probability of DR-UNet for the source domain is trained in a supervised manner. Then, we send the \textit{softmax} output simultaneously to output-space alignment and entropy minimisation discriminators. The domain classifier networks, $\mathbf{D_{1}}$ and $\mathbf{D_{2}}$, then differentiates whether its input is from source domain or target domain.} 
    \label{figdiagaram}
\end{figure*}

Our proposed UDA framework consists of a multi-task segmentation and point-cloud regression network $\mathbf{G}$, which has shared weights between  $\mathbb{D}_{s}$ and  $\mathbb{D}_{t}$. We modified DR-UNet~\cite{Vesal2019} segmentation network by adding another head to the encoder part. This head estimates the latent representation features of the encoder to a vector of size $300 \times 3$, which is the shape information of the input image in the form of point-cloud. We constructed three discriminators \textit{viz.} $\mathbf{D_{3}}$ for point-cloud surface shape alignment, $\mathbf{D_{1}}$ for segmentation output adaptation and $\mathbf{D_{2}}$, which receives self-weighted information map for entropy minimisation. 

During training, we first provide the images $\mathbf{I}_{s} \in \mathcal{R}^{h \times w \times 3}$ (with annotations) from source domain to the segmentation network in a supervised manner to optimise $\mathbf{G}$. We then take the images from the target domain and predict the segmentation output $\mathbf{S}_{t} \in \mathcal{R}^{h \times w \times 4}$. Once we have the output probability maps for both domains, we convert them to self-weighted information to compute entropy. Then, the network back-propagates gradients from $\mathbf{D}_{1}$, $\mathbf{D}_{2}$, and $\mathbf{D}_{3}$ to $\mathbf{G}$, which encourages $\mathbf{G}$ to optimise its weights w.r.t the segmentation labels from source and target domains. Fig.~\ref{figdiagaram} shows an overview of the proposed network. Our pipeline takes advantage of entropy minimisation and output space alignment, so we briefly review these methods before introducing our shape-aware point-cloud based UDA method.

\noindent\textbf{Multi-task Segmentation and Point-Cloud Regression Network:} Our segmentation network (Fig.~\ref{figSegNet}), $\mathbf{G}$, has an encoder that extracts high-level features from input images. The encoder has four levels and a bottleneck similar to DR-UNet~\cite{Vesal2019}, which also includes residual connections and dilated convolutions in the last level. We constructed two independent decoder heads for output segmentation and point-cloud regression. The decoder for segmentation is also similar to the DR-UNet network. It consists of four levels of 2D convolution, nearest neighbour up-sampling, and a $1\times1$ convolution. The point-cloud regression head has one convolution with a kernel size of $6\times6$ and a fully connected layer, which takes the encoder latent features and generates a shape vector (300$\times$3) of the point-cloud.

We use PatchGAN~\cite{DCGAN,Isola2018} for the discriminators $\mathbf{D}_{1}$ and $\mathbf{D}_{2}$, which takes two inputs (e.g. source and target probability maps or entropy maps) and distinguishes patches of size $8\times8$. PatchGAN discriminator mainly penalises structure at the scale of local image patches and is run convolutionally across the input images. There are five convolutional layers in this architecture, with a kernel size of $4\times4$ and a stride shape of 2,  and the number of feature maps for each layer is: $[64, 128, 256, 512, 1]$, respectively. Each convolution layer is followed by a leaky ReLU activation map, parameterised with 0.2.

\subsection{Output Feature Space Alignment} The first component of our network mainly focuses on adapting the distribution of output probability segmentation masks based on the adversarial learning approach similar to~\cite{TsaiCVPR2018}. The segmentation network $\mathbf{G}$ predicts output segmentation masks, and for adversarial learning, a discriminator is included to recognise whether the input image is from the source or target domains. Here, we denote images from source and target domains as $\mathbf{I_{s}}$, $\mathbf{I_{t}}$ $\in \mathcal{R}^{h\times w \times 3}$ . The segmentation network $\mathbf{G}$ first receives the source images $\mathbf{I_{s}}$ for supervised learning and predicts the segmentation output $\mathbf{S}_{s}$. Then we predict the segmentation output $\mathbf{S}_{t}$ for the target image $\mathbf{I_{t}}$ (without annotations). With an aim to align $\mathbf{S}_{s}$ and $\mathbf{S}_{t}$ closer to each other, $\mathbf{D_{1}}$ takes these predictions as its input to differentiate whether it is from $\mathbb{D}_{s}$ or $\mathbb{D}_{t}$. The network propagates gradients with an adversarial loss from $\mathbf{D_{1}}$ to $\mathbf{G}$ on the target prediction,  which would encourage $\mathbf{G}$ to produce more similar segmentation distributions in the target domain to the source prediction. The discriminator loss and adversarial loss can be defined as the following:

\begin{figure*}[ht]
    \centering
    \includegraphics[width=\textwidth]{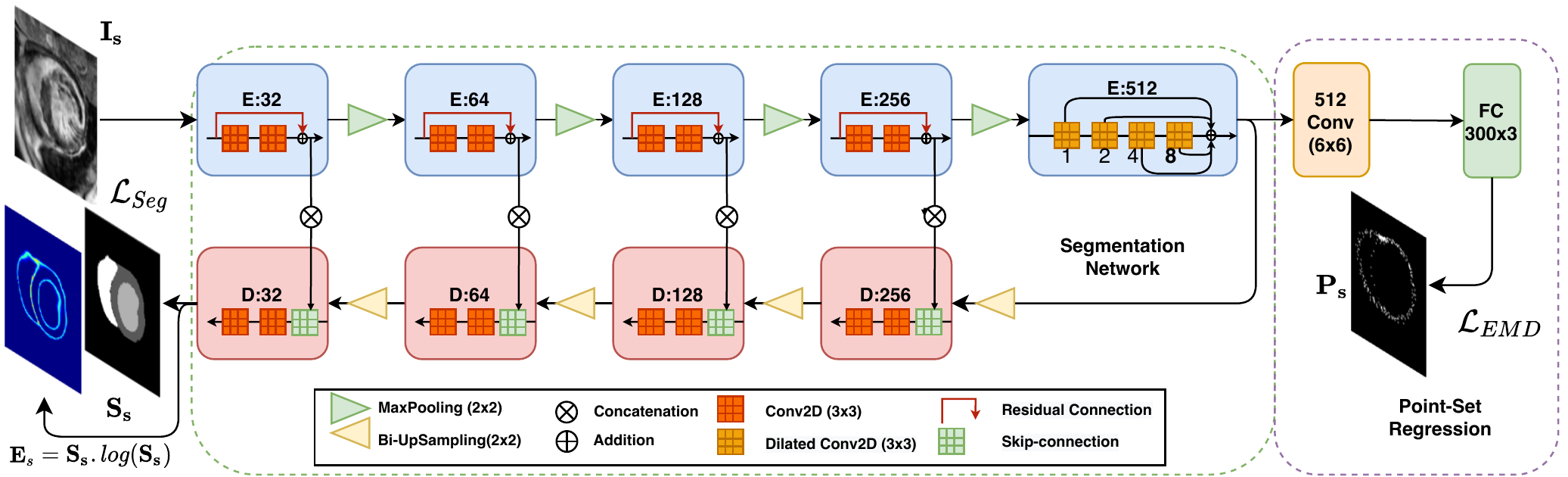}
    \caption{Multi-task segmentation and point-cloud regression network architecture. An input image from the source or target domain is fed into the model and the model predicts three outputs including, a segmentation probability map $\mathbf{S}_{s}$, an output entropy map $\mathbf{E}_{s}$ and a point-cloud vector $\mathbf{P}_{s}$, respectively.} 
    \label{figSegNet}
\end{figure*} 

\begin{equation}
     \mathcal{L}_{adv_{1}}(\mathbf{I}_{t}) =  \mathbb{E}_{x_{t}\sim \mathbf{I}_{t}} \log (\mathbf{1-D}_{1}(\mathbf{S}_{t})),  
     \label{eqnlv1}
\end{equation}
\begin{equation}
    \begin{split}
          \mathcal{L}_{D_{1}}(\mathbf{I}_{s}, \mathbf{I}_{t}) =  \mathbb{E}_{x_{t}\sim \mathbf{I}_{t}} \log (\mathbf{D}_{1}(\mathbf{S}_{t}))  + \\ \mathbb{E}_{x_{s}\sim \mathbf{I}_{s}} \log (\mathbf{1-D}_{1}(\mathbf{S}_{s})),  
    \end{split}
  \label{eqD1}
\end{equation}
Where $\mathbf{S}_{s}$ and $\mathbf{S}_{t}$ are the segmentation output from $\mathbb{D}_{s}$ and $\mathbb{D}_{t}$, respectively.

\subsection{Entropy Minimisation} Entropy minimisation showed great performance in the computer vision field for semantic segmentation. It is a regularisation method, which can prevent the model from overfitting and improve generalisation performance and robustness~\cite{vu2019advent, BEAL2019}. Entropy allows the exploration of the structural consistency between the two domains. For example, the cardiac LGE and cardiac CT images have no clear boundaries around the left ventricle (LV) and myocardium (Myo) , which makes it more difficult to segment in comparison to the bSSFP-MRI. Therefore, it enforces the segmentation network to generate high-entropy (uncertainty) on the soft boundary regions. To tackle the issue of uncertain predictions, we further adopt an entropy-driven adversarial learning model similar to~\cite{vu2019advent} to reduce the performance gap between the source and target domains by enforcing the entropy maps of the target domain predictions to be similar to the source ones. Given the pixel-wise mask probability prediction $S_{s}$ of input image $\mathbf{I}_{s}$, we use the Shannon Entropy to calculate the entropy map at pixel-level~\cite{vu2019advent} shown in Eq.~\ref{ent}:

\begin{equation}
    \mathcal{\mathbf{E}}_{s} (\mathbf{I}_{s}) = -\frac{1}{N}\sum_{n=1}^{N}\sum_{c=1}^{C}\mathbf{S}^{n,c}_{s} log(\mathbf{S}^{n,c}_{s})
    \label{ent}
\end{equation}

where $n$ indicates image number and $c$ indicates channel number. To conduct the entropy-driven adversarial learning, we constructed $\mathbf{D}_{2}$ as an entropy discriminator network to align the entropy maps between $\mathbf{E}_{s}$ and $\mathbf{E}_{t}$ and enforce the segmentation network to minimise the entropy in the target domain. Similar to output space alignment, the entropy discriminator aims to figure out whether the entropy map is from the source or the target domain.
\begin{equation}
    \mathcal{L}_{adv_{2}}(\mathbf{I}_{t}) =  \mathbb{E}_{x_{t}\sim \mathbf{I}_{t}} \log (\mathbf{1-D}_{2}(\mathbf{E}_{t})),  
    \label{eqnlv2}
\end{equation}
\begin{equation}
    \begin{split}
     \mathcal{L}_{D_{2}}(\mathbf{I}_{s}, \mathbf{I}_{t}) =  \mathbb{E}_{x_{t}\sim \mathbf{I}_{t}} \log (\mathbf{D}_{2}(\mathbf{E}_{t})) + \\ \mathbb{E}_{x_{s}\sim \mathbf{I}_{s}} \log (\mathbf{1-D}_{2}(\mathbf{E}_{s})),
    \end{split}
     \label{eqD2}
\end{equation}
Where $\mathbf{E}_{s}$ and $\mathbf{E}_{t}$ are the self-weighted information maps from $\mathbb{D}_{s}$ and $\mathbb{D}_{t}$, respectively~(cf.~Eq.~\ref{ent}).

\subsection{Point-cloud Shape Alignment} Incorporating shape information could improve the performance of a medical image segmentation model. One way to generate an overall shape of an anatomical object is using ``point-clouds". Point-clouds contain data that expresses the external surface of an object onto a three-dimensional structure using the euclidean $x, y, z$ coordinates. It is represented as a set of 3D points ${P_{i}| i = 1, ..., n}$, where each point $P_{i}$ is a vector of its $(x, y, z)$ coordinates.

Learning shape representation with point-clouds for target structures with distinct anatomy could be useful for the segmentation of target shapes. Hence, we developed a novel UDA using point-clouds shape learning combined with feature adaptation to improve the segmentation performance of our UDA. Our segmentation network has another head attached that takes deep encoded features containing object shape information in the form of point-clouds. Our model outputs shape representations of target cardiac structure; where the shape representations are sets of points located on the combined surface of LV, RV, and Myo. Our point-cloud network is inspired by point set generator (Fan~\emph{et al.}~\cite{FanCVPR2017}). Here the authors use a point-cloud generator which is built with 2D CNN layers. It takes 2D images and random vectors as its inputs and outputs sets of points as the 3D reconstruction of target objects. To obtain 3D reconstruction, a moderate level of uncertainty is desirable and useful, which Fan~\emph{et al.} achieve by the use of the random vectors~\cite{FanCVPR2017}. 

In our multi-task learning setup, the proposed point-cloud regressor is jointly optimised, with the DR-UNet segmentation model. Fig.~\ref{figSegNet} shows the pipeline of our multi-task learning network. This model directly consumes unordered point-clouds as inputs. The point-cloud regressor network takes the deep latent features extracted by DR-UNet encoder. It has a convolution layer with a kernel size $6 \times 6$ and a fully connected layer which predicts a vector of $300 \times 3$ that represent the point-clouds of the target image.  

\subsubsection{Point-cloud Ground-Truth Generation} To generate ground-truth point-clouds, we combined LV, RV, and Myo annotations to produce the external heart surface as a binary mask. The ground-truth surface points are generated using the marching cubes algorithm and farthest point sampling~\cite{FanCVPR2017}. The size of the point-cloud ($N_{P}$) is empirically set to 300 in all experiments, with coordinates normalised in the range of $(0,1)$. The regression loss is measured with Earth Mover’s Distance (EMD) metric~\cite{FanCVPR2017}. 

\subsubsection{Point-cloud Objective Function}
To compare the predicted point-cloud matrix and the ground-truth, we need a suitable loss function that includes the inherent uncertainty of the point-cloud predictions. There are several loss functions for point-cloud estimation such as the Hausdorff distance (HD) and Chamfer distance (CD)~\cite{FanCVPR2017}. However, these objective functions are not robust to the outliers. Here, to train our point-cloud regression network, we employed Earth Mover’s Distance (EMD)~\cite{FanCVPR2017}. Consider $\mathbf{P}_{s}$, $\mathbf{P}_{gt} \in \mathbb{R}^3{}$ of equal size $|\mathbf{P}_{s}| = |\mathbf{P}_{gt}|$, where $\mathbf{P}_{s}$ is the prediction and $\mathbf{P}_{gt}$ is the ground truth. The EMD loss between $\mathbf{P}_{s}$ and $\mathbf{P}_{gt}$ is defined as:
\begin{equation}
    \mathcal{L}_{EMD}(\mathbf{P}_{s}, \mathbf{P}_{gt}) = \argmin_{\theta :\mathbf{P}_{s}\rightarrow \textbf{P}_{gt} }\Sigma_{x \in \mathbf{P}_{s}}||x - \theta (x)||_{2} 
    \label{eqemd}
\end{equation}
where $\theta :\mathbf{P}_{s}\rightarrow \mathbf{P}_{gt}$ is a  one-to-one correspondence. The EMD tries to solve an optimisation problem called the assignment problem. For every but a zero measure subset of point set pairs, the optimal one-to-one correspondence $\theta$ is unique and invariant under the infinite flow of the points. Therefore, EMD is differentiable practically everywhere.

\subsection{Point-Net Discriminator}
\begin{figure}[ht]
    \centering
    \includegraphics[width=\columnwidth]{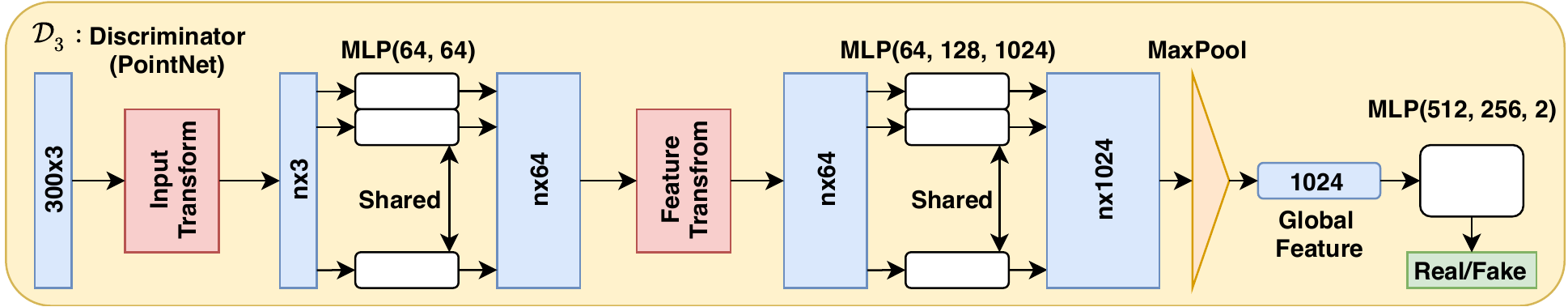}
    \caption{PointNet discriminator architecture. It takes an input size of $300 \times 3$ and applies a geometric transformation on input and feature level. The MaxPooling layer shrinks the features to generate global features, and there are three fully connected layers with a \textit{softmax} activation function that classify the point-cloud as real(source) or fake(target). } \label{figpointdisc}
\end{figure}

The main problem with point-clouds is that common convolutional architecture expects highly regular input data format, like the image or temporal features. To be able to adapt the point-cloud features between two domains, there is a need for a network which solely processes point-clouds. Inspired by ``PointNet" architecture~\cite{CharlesCVPR2017}, we propose a discriminator for adversarial adaptation of source and target domains based on point-clouds. This network learns a spatial encoding of every point within the point-clouds and then aggregate all the features to a global vector. Fig.~\ref{figpointdisc}. presents the overall architecture of our proposed PointNet discriminator network $\mathbf{D_{3}}$. The discriminator network receives 300 $\times$ 3 points as input and discriminates whether it is from source or target domain. In this network, we employ transformation layers proposed in~\cite{CharlesCVPR2017} to enable learned representation by the network to be invariant to geometric alterations. For all the layers, we used Batch Normalisation with the ReLU activation function. There are fully connected layers in the network that aggregate learned optimal values into the global descriptor for the complete shape. To train the discriminator binary cross-entropy loss is used for adversarial learning, which identifies the generated points $\mathbf{P}_{s}$ for source domain versus target domain $\mathbf{P}_{t}$. The adversarial loss and discriminator loss is similar to entropy and output space alignment, which is defined as follows:
\begin{equation}
   \mathcal{L}_{adv_{3}}(\mathbf{I}_{t}) =  \mathbb{E}_{x_{t}\sim \mathbf{I}_{t}} \log (\mathbf{1-D}_{3}(\mathbf{P}_{t})),  
   \label{eqnlv3}
\end{equation}
\begin{equation}
    \begin{split}
        \mathcal{L}_{D_{3}}(\mathbf{I}_{s}, \mathbf{I}_{t}) =  \mathbb{E}_{x_{t}\sim \mathbf{I}_{t}} \log (\mathbf{D}_{3}(\mathbf{P}_{t})) +  \\ \mathbb{E}_{x_{s}\sim \mathbf{I}_{s}} \log (\mathbf{1-D}_{3}(\mathbf{P}_{s})),
    \end{split}
\label{eqD3}
\end{equation}

\noindent\textbf{Total Objective Function:} To train the proposed method in an end-to-end fashion,  we use the binary cross-entropy + Jaccard loss ($\mathcal{L}_{seg}$)~\cite{wang2019boundary} as the segmentation loss for supervised training of source domain. To obtain the total loss, we combine Eq. \ref{eqnlv1}, \ref{eqD1}, \ref{eqnlv2}, \ref{eqD2}, \ref{eqemd}, \ref{eqnlv3} and \ref{eqD3} which can be formulated as below. 
\begin{equation}
    \begin{split}
        \mathcal{L}_{total}(\mathbf{I}_{s}, \mathbf{I}_{t}) = \underbrace{\mathcal{L}_{seg}(\mathbf{I}_{s}) + \mathcal{L}_{EMD}(\mathbf{I}_{s})}_\text{supervised} + \\ \underbrace{\lambda_{adv_{1}}\mathcal{L}_{adv1}(\mathbf{I}_{t}) - \lambda_{D_{1}}\mathcal{L}_{D_{1}}(\mathbf{I}_{s}, \mathbf{I}_{t})}_\text{output space} + \\ \underbrace{\lambda_{adv_{2}}\mathcal{L}_{adv2}(\mathbf{I}_{t}) - \lambda_{D_{2}}\mathcal{L}_{D_{2}}(\mathbf{I}_{s}, \mathbf{I}_{t})}_\text{entropy space} + \\ \underbrace{\lambda_{adv_{3}}\mathcal{L}_{adv3}(\mathbf{I}_{t}) - \lambda_{D_{3}}\mathcal{L}_{D_{3}}(\mathbf{I}_{s}, \mathbf{I}_{t})}_\text{Point-cloud space}
    \end{split}
\end{equation}

where $\lambda_{adv_{1}}, \lambda_{D_{1}}, \lambda_{adv_{2}}, \lambda_{D_{2}}, \lambda_{adv_{3}}, \lambda_{D_{3}}$ are weights to balance the losses. We empirically set these values to 1 for adversarial losses and 0.2 for discriminator losses. The overall min-max optimisation problem can be summarised as following:
\begin{equation}
    \argmin_{\mathbb{D}_s, \mathbb{D}_t}\argmax_{\mathbf{D}_{1},\mathbf{D}_{2}, \mathbf{D}_{3}} \mathcal{L}_{total}(\mathbf{I}_{s}, \mathbf{I}_{t})
\end{equation}
where $\mathbb{D}_s$ and $\mathbb{D}_t$ are networks with shared trainable weights.

\section{Experimental Results}
\subsection{Datasets}
To validate our proposed method for UDA segmentation, we utilised two datasets.

\subsubsection{Cardiac Multi-sequence Segmentation.} This STACOM MS-CMRSeg~\cite{zhuang2020cardiac} 2019 challenge dataset  contains 45 short-axis Cine-MRI scans from patients diagnosed with cardiomyopathy. The dataset was collected in Shanghai Renji hospital and institutional ethics approved for all patients data~\cite{zhuang2}. For each patient, three different Cine-MR sequences were acquired, \textit{viz.} bSSFP, LGE, and T2-weighted. The ground-truth contours were generated by two experts and include the right ventricle (RV) cavity, the LV cavity, and the myocardium region. The LGE-MRI is a T1-weighted sequence with inversion-recovery, consisting of $10-18$ slices comprising the main section of the ventricles. These images have a resolution of $512\times512$ pixels, with an in-plane resolution of $0.75\times0.75$ mm and a slice thickness of 5 mm. The bSSFP-MRI sequences consist of $8-12$ slices, including the full ventricles from the basal plane of the mitral valve to the apex. These have an image resolution of $1.25\times1.25$ mm and a slice thickness of $8-13$ mm. The T2-weighted-MRI sequences have fewer number of slices between $3-7$, with an in-plane resolution of $1.35\times1.35$ mm and slice thickness is $12-20$ mm. Since T2-weighted and bSSFP sequences had very few slices, we combined both (henceforth called bSSFP-CMR) as the source domain and used LGE-MRI as the target domain. The images within MS-CMRSeg 2019 dataset has a high level of variation in contrast and brightness. The variability results from different system settings and data acquisition, which makes it harder for neural networks to process the images. Therefore, we enhanced the cine-MR image contrast slice wise, using histogram equalisation. The MR sequences are normalised using min-max normalisation and centre-cropped to $224\times224$ pixels to have only region-of-interest (ROIs) areas. However, we found this is not an optimal solution because cardiac anatomy can exist in different regions of MRI scan. Therefore, we have also implemented an ROI detector. The ROI detector is a shallow UNet\cite{Unett} that takes the binary mask of original images and predicts a coarse segmentation of RVs and LVs. Then we find the centre of the predicted segmentation mask and crop the real images with a size of $224\times224$ pixels based on this reference.

\subsubsection{Cardiac Cross-modality Segmentation.} To further evaluate our proposed method, we employed the Multi-Modality Whole Heart Segmentation (MM-WHS) Challenge 2017 dataset for cardiac segmentation \cite{MMWHS}. This dataset contains 20 MRI and 20 CT unpaired volumes with ground truth masks. For evaluating the domain adaptation on cardiac segmentation, we include the following four structures: ascending aorta (AA), the left atrium blood cavity (LA), the LV blood cavity, and the myocardium of the LV. We used the same data split as in \cite{dou2019pnp} for training (16 subjects) and testing (4 subjects) subsets in experiments. The dataset is prepossessed such that the MRI and CT images are re-oriented, resized and cropped centring at the heart region, such that the view of multi-modal images are roughly on the same page. We extract MRI and CT scans by 2D slices of size $256 \times 256$ at coronal plane during training and the whole dataset is normalised in 3D for each modality respectively. Data augmentations like rotation, zoom and affine transformation are utilised during the training.

\subsubsection{Evaluation Metrics} We employed commonly-used metrics, the Dice similarity coefficient (Dice), the Hausdorff distance (HD) \cite{zhuang2020cardiac} and Average surface distance (ASD) \cite{dou2019pnp}, to quantitatively evaluate the segmentation performance of models. Dice measures the voxel-wise segmentation accuracy between the predicted and reference volumes. HD and ASD calculate the maximum and average distances between the surface of the prediction mask and the ground-truth in 3D. Hence, a higher Dice value and a lower HD and ASD values indicate better segmentation. The evaluation is performed on the subject-level segmentation volume, to be consistent with the MS-CMRSeg challenge benchmark study as well as previous works~\cite{Chen2019UnsupervisedMS, Vesal2019, zhuang2019domain, dou2019pnp, chen2020unsupervised}.

\subsubsection{Implementation details}  We implemented our framework with the PyTorch 1.4 deep learning library. We trained the whole pipeline directly without any warm-up phase of supervised learning with a mini-batch of size $16$. The discriminators $\mathbf{D}_{1}$, $\mathbf{D}_{2}$, and $\mathbf{D}_{3}$,  were optimised with the SGD algorithm, while the Adam optimiser is utilised for the segmentation network $\mathbf{G}$. For MS-CMRSeg dataset, we set the initial learning rate of Adam as $0.001$ and reduced it by a factor of 0.2 every 100 epochs for a total of 600 epochs. The learning rate of discriminator training was set to $0.000025$. 

For MM-WHS dataset, the learning rate for the segmentation network $\mathbf{G}$ is set to 0.0002 without any learning rate decay. Compared with the learning rate setting of the MS-CMRSeg dataset, we have more meticulous configuration for the discriminators and the corresponding adversarial learning for the MM-WHS dataset. The models were trained from scratch until there is no further improvement for the last 100 epochs. We employed the same preprocessing and normalisation scheme as in \cite{dou2019pnp}. For a fair comparison with other methods, we do not apply any extra data augmentations.

 \begin{table*}[htbp]
   \centering
   \caption{Performance comparison between our proposed method and different unsupervised and supervised domain adaption methods for cardiac multi-modal segmentation based on volumetric Dice and HD. The values are shown with $(mean \pm std)$.}
     \resizebox{\textwidth}{!}{
     \begin{tabular}{lccccccccccccccccl}
      \hline
     Methods & \multicolumn{8}{c|}{Volumetric Dice $[mean \pm std]$ $\uparrow$}  & \multicolumn{8}{c|}{Volumetric HD [mm] $\downarrow$} & \multicolumn{1}{c}{Training} \\
    \cline{2-17}        & \multicolumn{2}{c}{Myo} & \multicolumn{2}{c}{LV} & \multicolumn{2}{c}{RV} & \multicolumn{2}{c|}{Average Dice} & \multicolumn{2}{c}{Myo} & \multicolumn{2}{c}{LV} & \multicolumn{2}{c}{RV} & \multicolumn{2}{c|}{Average HD} \\
    \hline
    \hline
     Baseline (W/o DA) & \multicolumn{2}{c}{0.392$\pm$0.210} & \multicolumn{2}{c}{0.651$\pm$0.265} & \multicolumn{2}{c}{0.619$\pm$0.270} & \multicolumn{2}{c|}{0.554$\pm$0.248} & \multicolumn{2}{c}{20.54$\pm$7.582} & \multicolumn{2}{c}{15.88$\pm$10.723} & \multicolumn{2}{c}{22.32$\pm$17.999} & \multicolumn{2}{c|}{19.59$\pm$12.101} & Unsupervised\\
    
     \hline
     Chen et al.~\cite{Chen2019UnsupervisedMS} & \multicolumn{2}{c}{\textbf{0.826$\pm$0.035}} & \multicolumn{2}{c}{\textbf{0.919$\pm$0.026}} & \multicolumn{2}{c}{0.875$\pm$0.050} & \multicolumn{2}{c|}{\textbf{0.873$\pm$0.037}} & \multicolumn{2}{c}{10.28$\pm$3.376} & \multicolumn{2}{c}{12.45$\pm$3.142} & \multicolumn{2}{c}{15.38$\pm$6.942} & \multicolumn{2}{c|}{12.703$\pm$4.487} & Unsupervised\\
     \textbf{Proposed method} & \multicolumn{2}{c}{0.794$\pm$0.041} & \multicolumn{2}{c}{0.909$\pm$0.032} & \multicolumn{2}{c}{\textbf{0.878$\pm$0.053}} & \multicolumn{2}{c|}{0.860$\pm$0.042} & \multicolumn{2}{c}{\textbf{9.390$\pm$2.628}} & \multicolumn{2}{c}{\textbf{7.647$\pm$3.231}} & \multicolumn{2}{c}{\textbf{8.452$\pm$2.831}} & \multicolumn{2}{c|}{\textbf{8.496$\pm$2.897}} & \textbf{Unsupervised} \\

     ADVENT~\cite{vu2019advent} & \multicolumn{2}{c}{0.778$\pm$0.061} & \multicolumn{2}{c}{0.906$\pm$0.034} & \multicolumn{2}{c}{0.867$\pm$0.063} & \multicolumn{2}{c|}{0.850$\pm$0.053} & \multicolumn{2}{c}{9.727$\pm$2.686} & \multicolumn{2}{c}{7.375$\pm$3.311} & \multicolumn{2}{c}{9.952$\pm$3.459} & \multicolumn{2}{c|}{9.018$\pm$3.152} & Unsupervised \\
     Wang et al.~\cite{Wang2019} & \multicolumn{2}{c}{0.796$\pm$0.059} & \multicolumn{2}{c}{0.896$\pm$0.047} & \multicolumn{2}{c}{0.846$\pm$0.086} & \multicolumn{2}{c|}{0.846$\pm$0.064} & \multicolumn{2}{c}{13.59$\pm$5.206} & \multicolumn{2}{c}{15.7$\pm$5.814} & \multicolumn{2}{c}{15.21$\pm$6.327} & \multicolumn{2}{c|}{14.833$\pm$5.782} & Unsupervised \\
     Ly et al.~\cite{Ly} & \multicolumn{2}{c}{0.705$\pm$0.115} & \multicolumn{2}{c}{0.87$\pm$0.051} & \multicolumn{2}{c}{0.762$\pm$0.150} & \multicolumn{2}{c|}{0.779$\pm$0.105} & \multicolumn{2}{c}{41.74$\pm$7.696} & \multicolumn{2}{c}{42.79$\pm$13.26} & \multicolumn{2}{c}{34.38$\pm$8.065} & \multicolumn{2}{c|}{39.637$\pm$9.674} & Unsupervised \\
     \hline
     Wang et al.~\cite{Wang22019} & \multicolumn{2}{c}{0.843$\pm$0.048} & \multicolumn{2}{c}{0.926$\pm$0.028} & \multicolumn{2}{c}{0.890$\pm$0.044} & \multicolumn{2}{c|}{0.886$\pm$0.04} & \multicolumn{2}{c}{9.748$\pm$3.28} & \multicolumn{2}{c}{11.65$\pm$4.002} & \multicolumn{2}{c}{13.34$\pm$4.615} & \multicolumn{2}{c|}{11.579$\pm$3.966} & Supervised \\
     Campello et al.~\cite{Campello2019} & \multicolumn{2}{c}{0.81$\pm$0.061} & \multicolumn{2}{c}{0.898$\pm$0.045} & \multicolumn{2}{c}{0.866$\pm$0.050} & \multicolumn{2}{c|}{0.858$\pm$0.052} & \multicolumn{2}{c}{10.78$\pm$4.066} & \multicolumn{2}{c}{11.96$\pm$3.62} & \multicolumn{2}{c}{15.91$\pm$6.895} & \multicolumn{2}{c|}{12.883$\pm$4.86} & Supervised \\
     Vesal et al.~\cite{Vesal2019} & \multicolumn{2}{c}{0.789$\pm$0.073} & \multicolumn{2}{c}{0.912$\pm$0.034} & \multicolumn{2}{c}{0.833$\pm$0.084} & \multicolumn{2}{c|}{0.845$\pm$0.064} & \multicolumn{2}{c}{11.29$\pm$4.559} & \multicolumn{2}{c}{12.54$\pm$3.379} & \multicolumn{2}{c}{17.11$\pm$6.141} & \multicolumn{2}{c|}{13.647$\pm$4.693} & Supervised \\
     Roth et al.~\cite{Roth2019} & \multicolumn{2}{c}{0.78$\pm$0.047} & \multicolumn{2}{c}{0.89$\pm$0.043} & \multicolumn{2}{c}{0.844$\pm$0.063} & \multicolumn{2}{c|}{0.838$\pm$0.051} & \multicolumn{2}{c}{11.58$\pm$7.524} & \multicolumn{2}{c}{16.25$\pm$6.336} & \multicolumn{2}{c}{18.12$\pm$9.262} & \multicolumn{2}{c|}{15.317$\pm$7.707} & Supervised \\
     Liu et al.~\cite{Liu2019} & \multicolumn{2}{c}{0.751$\pm$0.119} & \multicolumn{2}{c}{0.884$\pm$0.07} & \multicolumn{2}{c}{0.791$\pm$0.165} & \multicolumn{2}{c|}{0.809$\pm$0.118} & \multicolumn{2}{c}{14.30$\pm$8.17} & \multicolumn{2}{c}{14.75$\pm$7.823} & \multicolumn{2}{c}{17.87$\pm$9.322} & \multicolumn{2}{c|}{15.64$\pm$8.438} & Supervised \\
     Chen et al.~\cite{Chen22019} & \multicolumn{2}{c}{0.61$\pm$0.102} & \multicolumn{2}{c}{0.824$\pm$0.068} & \multicolumn{2}{c}{0.71$\pm$0.135} & \multicolumn{2}{c|}{0.715$\pm$0.102} & \multicolumn{2}{c}{23.69$\pm$14.66} & \multicolumn{2}{c}{24.62$\pm$12.66} & \multicolumn{2}{c}{23.46$\pm$7.596} & \multicolumn{2}{c|}{23.923$\pm$11.639} & Supervised \\
     \hline
     \hline
     Inter-Observer~\cite{zhuang2020cardiac} & \multicolumn{2}{c}{0.764$\pm$0.069} & \multicolumn{2}{c}{0.881$\pm$0.064} & \multicolumn{2}{c}{0.816$\pm$0.084} & \multicolumn{2}{c|}{0.82$\pm$0.072} & \multicolumn{2}{c}{12.03$\pm$4.443} & \multicolumn{2}{c}{14.32$\pm$5.164} & \multicolumn{2}{c}{21.53$\pm$9.46} & \multicolumn{2}{c|}{15.96$\pm$6.356} & Inter-Ob \\
     \hline
     \end{tabular}}
   \label{tab:addlabel}
 \end{table*}
 
\begin{table*}[htbp]
  \centering
  \caption{Performance comparison between our proposed method and different UDA methods for cardiac cross-modality segmentation (MRI $\rightarrow$ CT) based on volumetric Dice and ASD. The values are shown with $(mean)$.}
    \begin{tabular}{lccccc|ccccc}
    \hline
    Methods & \multicolumn{5}{c|}{Volumetric Dice $[mean \pm std]$ $\uparrow$} & \multicolumn{5}{c}{Volumetric ASD [mm] $\downarrow$} \\
\cline{2-11}          & AA    & LA    & LV    & Myo   & Average Dice & AA    & LA    & LV    & Myo   & Average HD \\
    \hline
    \hline
    Baseline (W/o DA) & 0.303 & 0.846 & 0.360 & 0.517 & 0.507 & 18.74 & 2.30  & 13.10 & 7.43  & 10.39 \\
    Baseline (W/o DA) + Point-Cloud & 0.787 & 0.744 & 0.315 & 0.335 & 0.545 & 3.94  & 2.94  & 43.18 & 12.00 & 15.51 \\
    \hline
    CycleGAN \cite{Zhu2017UnpairedCYCLEGAN} & 0.738 & 0.757 & 0.523 & 0.287 & 0.576 & 11.50 & 13.60 & 9.20  & 8.80  & 10.80 \\
    ADVENT \cite{vu2019advent} & 0.812 & 0.765 & 0.329 & 0.225 & 0.533 & 3.68  & 3.63  & 15.41 & 28.71 & 12.86 \\
    PnP-AdaNet \cite{dou2019pnp} & 0.740 & 0.689 & 0.619 & 0.508 & 0.639 & 12.80 & 6.30  & 17.40 & 14.70 & 12.80 \\
    SIFA \cite{chen2019synergistic} & 0.811 & 0.764 & \textbf{0.757} & \textbf{0.587} & \textbf{0.730} & 10.60 & 7.40  & 6.70  & 7.80  & 8.10 \\
    Proposed Method & \textbf{0.830} & \textbf{0.813} & 0.672 & 0.584 & 0.725 & \textbf{2.90} & \textbf{2.66} & \textbf{6.32} & \textbf{6.38} & \textbf{4.56} \\
    \hline
    \end{tabular}%
  \label{tab:ctmrtab1}%
\end{table*}%

\subsection{Comparison with other methods} 
To demonstrate the effectiveness of our proposed UDA method for leveraging multi-modal data, we compare it with both supervised and unsupervised learning methods. 

\textbf{MS-CMRSeg Dataset}: Table \ref{tab:addlabel} shows the quantitative results of different algorithms for the MS-CMRSeg challenge, specifically for the 40 LGE-MRI segmentation test data. We first compare with the model trained with only limited labelled LGE-MRI data $\mathbb{D}_{t}$ (referred as Supervised only) and take all other methods (Unsupervised, Inter-Ob) for benchmark study including ADVENT~\cite{vu2019advent} and AdaptSeg~\cite{TsaiCVPR2018} for comparison. Besides two-stage approaches, we also compare with the end-to-end model and inter-observer study. Meanwhile, we also compare with segmentation methods like Chen~\etal~\cite{Chen2019UnsupervisedMS} (Unsupervised) and  Wang~\etal~\cite{Wang22019} (Supervised), that achieved SOTA performance in MS-CMRSeg~\cite{zhuang2020cardiac} cardiac challenge segmentation. Supervised methods in MS-CMRSeg challenge had access to only 5 LGE-MRI studies to train their models where 2 or 3 studies are used for training and cross-validation.

As shown in Table~\ref{tab:addlabel}, the baseline method without DA performed poorly, which emphasises the domain shift between the source (bSSFP-MRI) and the target domain (LGE-MRI). The best model with supervised training regime achieved $88.6\%$ mean Dice and $11.57$ HD score by taking only limited labelled target data (5 LGE-MRI subjects) during training. When two types of data sources are available, unsupervised methods achieve comparable Dice scores to the supervised-only model by utilising adversarial training or generating synthetic data. Compared with the supervised-only methods, Chen~\etal~\cite{Chen2019UnsupervisedMS} approach further improved the segmentation performance, demonstrating the effectiveness of leveraging multi-sequence data $\mathbb{D}_{s}$ and unlabelled data $\mathbb{D}_{t}$ for training. 

\begin{figure*}[htbp]
\centering
    \includegraphics[width=17.4cm]{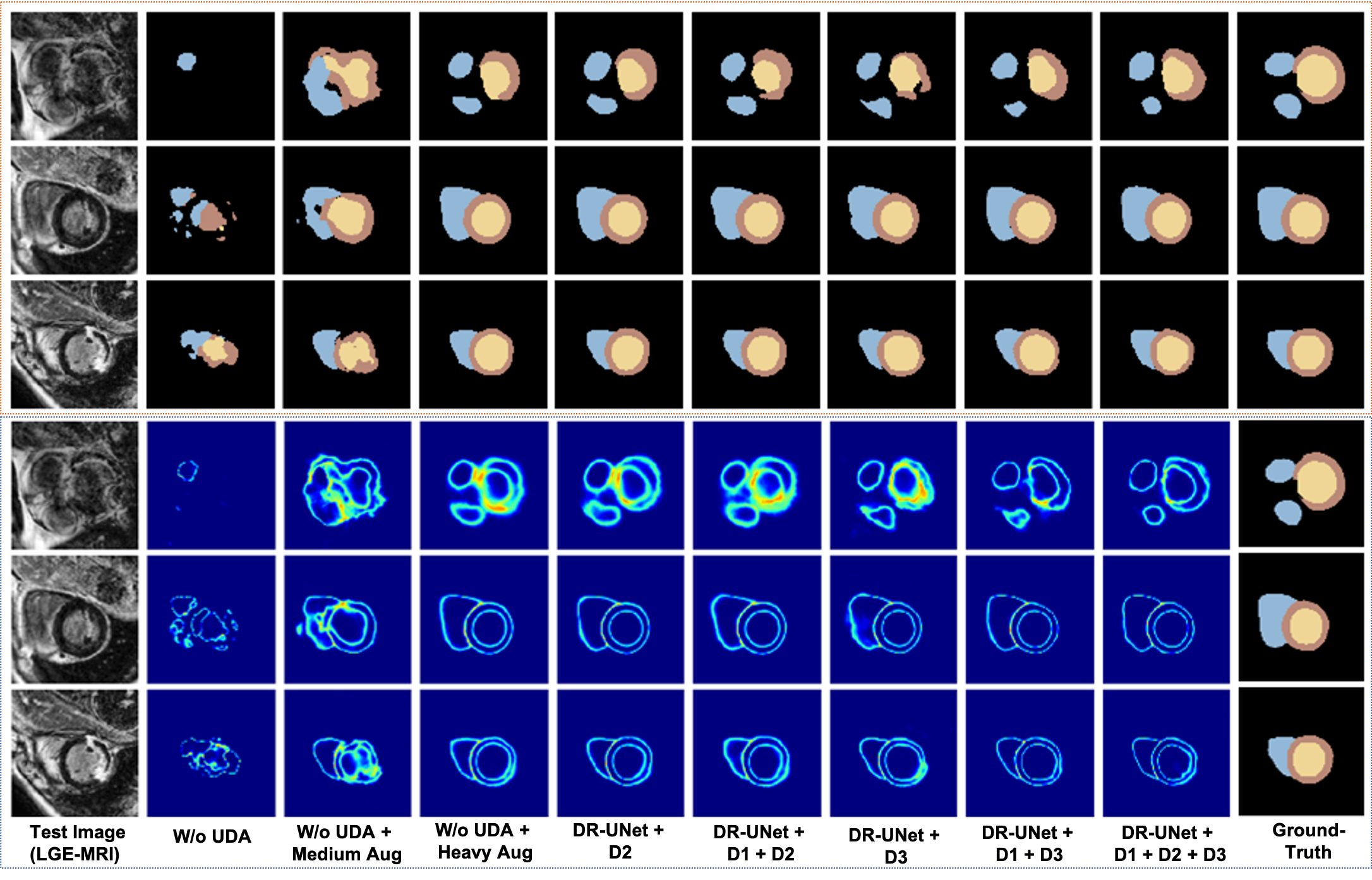}
    \caption{A visual comparison of segmentation output and entropy map results produced by different methods for LGE images. From left to right are the raw LGE test images (1st column), W/o Adaptation (\nth{2} column), W/o Adaptation but with medium and high-level augmentation (\nth{3}~\&~\nth{4}~column), results of our UDA segmentation methods (\nth{5}-\nth{9} column), and ground-truth (last column). The LV, Myo, and RV are indicated, in yellow, brown, and blue colour, respectively. Each row corresponds to one example.} \label{figoutput}
    \vspace{-1.0em}
\end{figure*}

The volumetric Dice scores of our proposed method are $0.794\pm0.041$ (Myo), $0.909\pm0.032$ (LV), and $0.878 \pm 0.053$ (RV), respectively and the volumetric HD values are respectively $9.390\pm2.628 mm$ (Myo), $7.647\pm3.231mm$  (LV), and $8.452\pm2.831mm$ (RV). The average Dice score of our method ranks second in the Table~\ref{tab:addlabel}, and it is 4 points higher as compared to the inter-observer Dice scores. Our method achieved the lowest average HD score in comparison to all existing approaches. The main reason can be in two folds. First, our network employs point-cloud to incorporate shape information during training which enforces explicitly in producing a more smooth surface and accurate segmentation output. Secondly, the concurrent adaptation in different spaces could lead to a better model optimisation. Interestingly, the unsupervised methods performed comparably to supervised ones. Specifically, LV structure has a different geometry shape from base to apex, but it is simpler to segment and has higher Dice scores in comparison to RV and Myo in most of the methods. We reported the slice-wise accuracy of different methods for different positions in the discussion section (Table~\ref{tab:slicewise}). The myocardial wall is typically the challenging structure to segment, especially in this dataset, in which some of the cases in the test data included scars. In comparison to the inter-observer study, our method achieved higher Dice scores for RV and Myo with 6.0\% and 3.0\% improvements, respectively. Fig.~\ref{figoutput} illustrates some qualitative examples comparing our proposed method with other SOTA approaches along with the baseline. Our model shows better results on LV classes, while ADVENT sometimes makes mistakes of predicting myocardial wall (\nth{1}~row,~\nth{5}~column). It is observed that our method better identifies heart substructures with a clean and accurate boundary, and produces less false positive predictions and more similar results to the ground-truth (\nth{9}~\&~\nth{10}~columns) compared with other methods. 

\textbf{MM-WHS Dataset}: Table \ref{tab:ctmrtab1} shows the quantitative results of different algorithms for the MM-WHS challenge dataset, specifically for the 4 CT segmentation test data. Similar to the previous dataset, we have compared our model with the baseline model without domain adaptation, feature-based and image-based UDA  segmentation  methods, including CycleGAN \cite{Zhu2017UnpairedCYCLEGAN}, ADVENT \cite{vu2019advent, dou2019pnp} and SIFA \cite{chen2019synergistic} that achieved SOTA performance in MM-WHS~\cite{MMWHS} benchmark dataset. For this dataset, we considered ASD metric and not HD, since almost all the existing methods reported their model performance with ASD.  As a baseline study, we obtained the ``Baseline W/o DA” lower bound by directly applying the model learned in MR source domain to target CT images without using any domain adaptation method. The poor performance of the baseline model indicates that the domain shift between the source and the target domain is significant. The baseline method after adding the point-cloud head improved the average volumetric Dice score to $54.5\%$, but produced a very high average ASD of $15.51 mm$. 

\begin{figure*}[ht]
    \centering
    \includegraphics[width=0.96\linewidth]{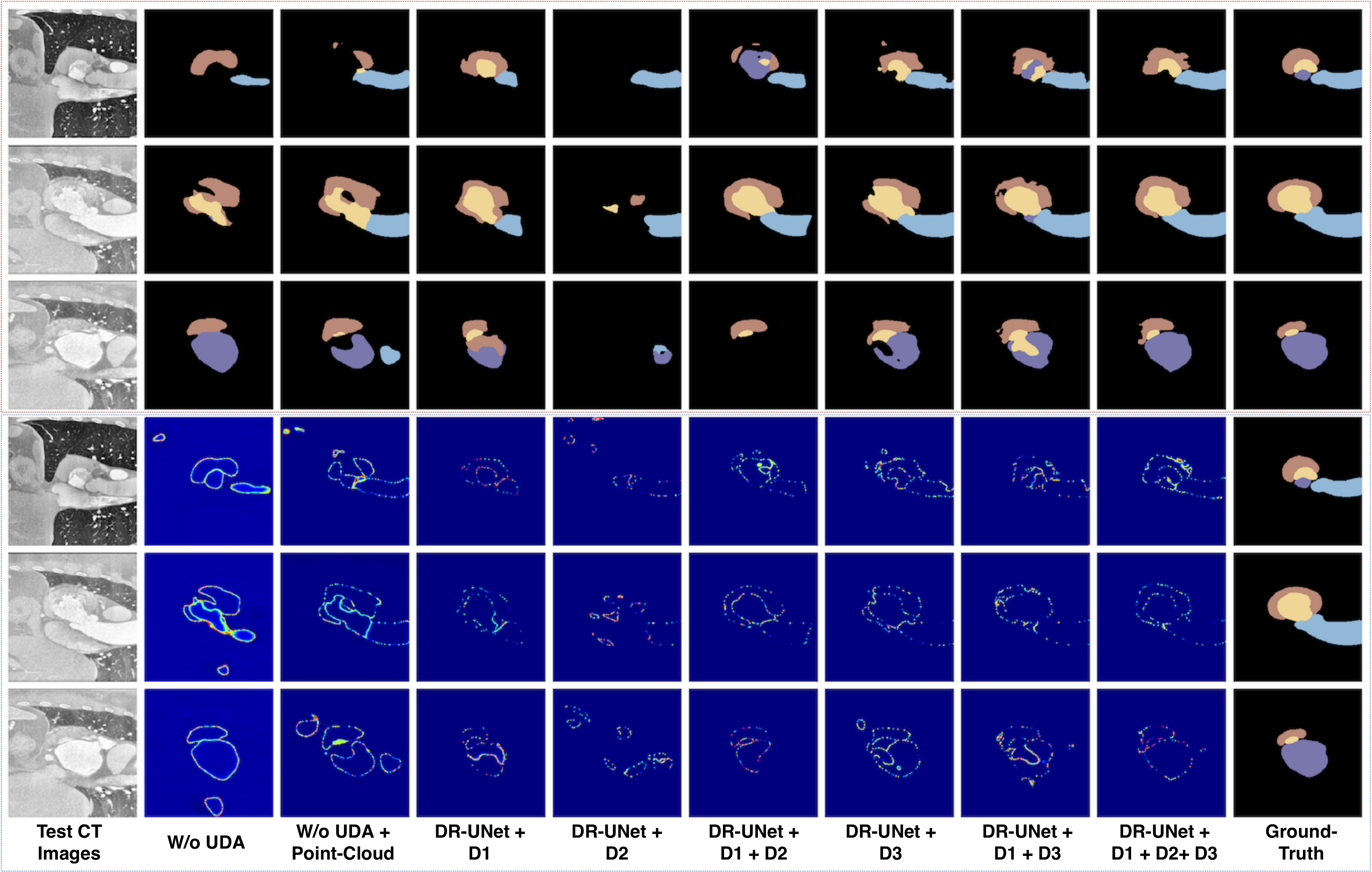}
    \caption{A visual comparison of segmentation output and entropy map results produced by different methods for CT images. From left to right are the raw CT test images (\nth{1} column), W/o Adaptation (\nth{2} column), W/o Adaptation with point-cloud as a multi-task(\nth{3}~column), results of our UDA segmentation methods (\nth{4}-\nth{9} column), and ground-truth (last column). The cardiac structures of AA, LA, LV, and Myo are indicated in blue, purple, yellow, and brown color respectively.} \label{fig:mrct_seg_etp}
    \vspace{-1.0em}
\end{figure*}

Remarkably, with our proposed network, the average Dice score improved to $72.5 \%$, and the ASD score reduced to $4.56 mm$. We achieved $83.0\%, 81.3\%, 67.2\%$ Dice scores for AA structure, LA and LV blood pools respectively. Notably, compared to SIFA model, which conducts both image and feature adaptations, our method achieved superior performance especially for the AA and LA structures, which have limited contrast in CT images. However, it produced a slightly worse segmentation output for Myo. The proposed network outperforms CycleGAN, ADVENT and PnP-AdaNet for segmentation of all cardiac structures. Fig.~\ref{fig:mrct_seg_etp} shows some qualitative examples comparing our proposed method with other SOTA approaches along with the baseline model for MM-WHS CT test images.  As it can be seen in the $\nth{3}$ row in Fig.~\ref{fig:mrct_seg_etp}, the LV and Myo structures have very limited intensity contrast with their surrounding tissues, but our method can make good predictions while all the other methods fail in this challenging case. Similar to MS-CMRSeg dataset, our model achieved the lowest average ASD score among all the networks. Hence, the point-cloud adaptation enforced the model to avoid over-segmentation of the cardiac structure and produced more smooth surface shape.

\subsection{Ablation study}
To evaluate whether adapting features in different spaces is truly beneficial for accurate cardiac segmentation on target modality, we performed an ablation study on the MS-CMRSeg benchmark dataset. First, we evaluate the effectiveness of data augmentation on the target domain. We test three different data augmentations strategies before implementing UDA. In the first attempt, we trained the segmentation network without any data augmentation, and the model achieved a mean volumetric Dice score of $0.554\pm0.248$ and an HD value of $19.584\pm12.101 mm$ (\nth{1}~data row, Table ~\ref{tab:addlabelAb}) on 40 LGE-MRI subjects of the target domain. In the next step, we applied histogram equalisation and rotate the myocardium with two angles, [$30^{\circ}, 60^{\circ}$]~\cite{Campello2019}, to increase the number of training samples. With this augmentation, we observed a $16.0\%$ improvement in mean volumetric Dice and minor improvement in volumetric HD (\nth{2}~data row, Table~\ref{tab:addlabelAb}). The reason for high volumetric HD values is mainly because the model could not learn the noisy border of LGE-images, which also has low contrast in those regions. Ultimately, we utilised an online imaging library (imgAug~\cite{imgaug}) and employed a set of extreme and complex augmentation such as affine-transformation, translation, gaussian noise, elastic deformation, contrast enhancement, optical deformation, etc. The segmentation network under these settings achieved a volumetric Dice score of $0.833\pm0.063$ and volumetric HD drastically reduced to $10.027\pm3.985 mm$ (\nth{3}~data row, Table~\ref{tab:addlabelAb}). 

\begin{figure}[t!]
    \centering
    \includegraphics[width=\linewidth]{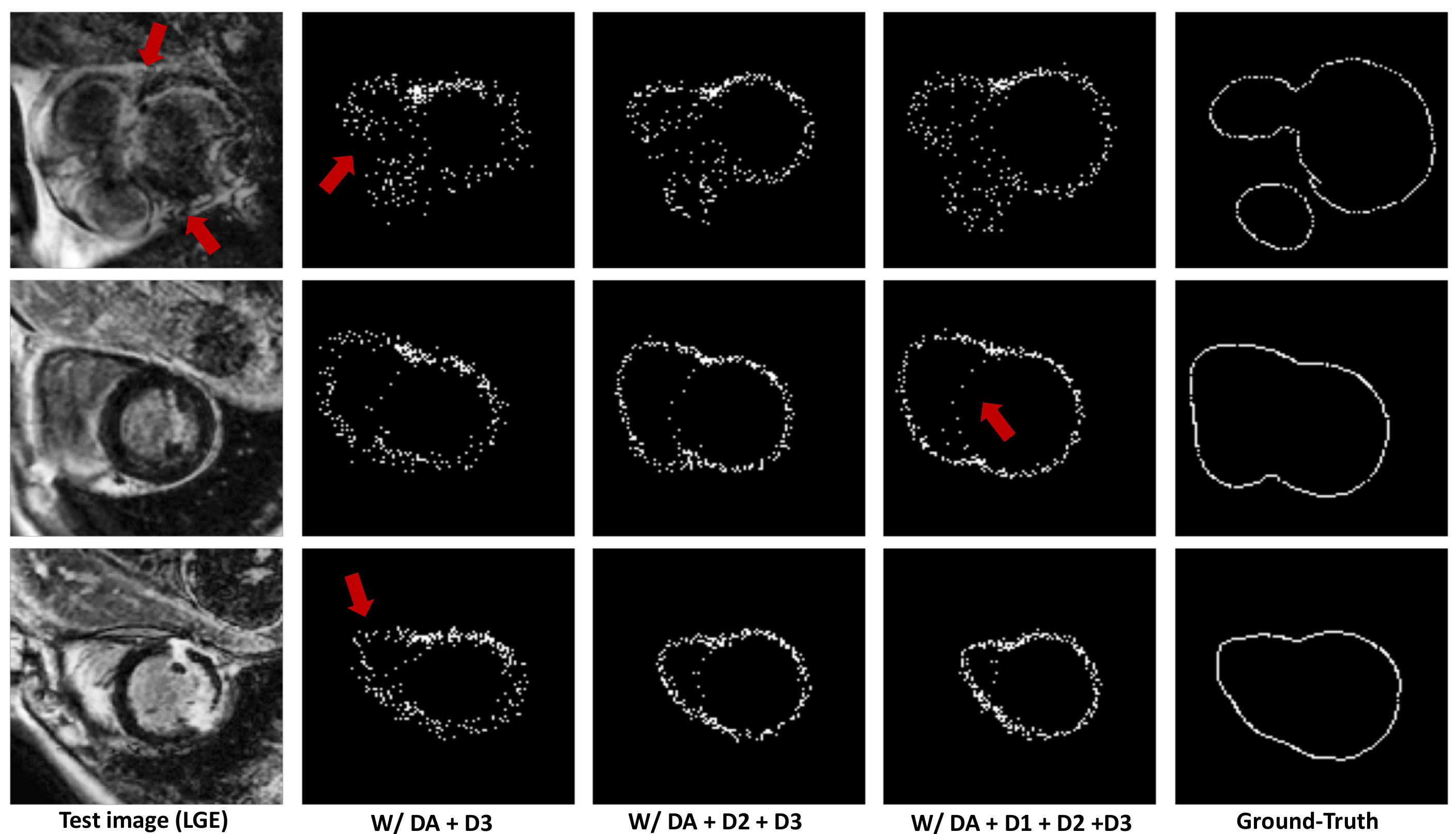}
    \caption{Visual comparison of point-clouds regression results produced by different methods for test LGE-MR images. From left to right are the raw LGE test images (\nth{1} column) and the results of our UDA methods (\nth{2}-\nth{4} column), and ground-truth point-cloud (last column). The red arrows point at the regions reconstructed by the model because of the noise in the image.} \label{figPoint_cloud}
    \vspace{-1.0em}
\end{figure}

Furthermore, we demonstrate that minimising the entropy in the target domain and adapting the shape using point-clouds can work jointly to improve domain adaptation performance. The quantitative and visual experimental analysis are shown in Table~\ref{tab:addlabelAb} and Fig.~\ref{figoutput} respectively. Our baseline network uses entropy minimisation only, which is constructed by removing the output space alignment and point-cloud adversarial loss when training the network. Compared with the “W/o UDA” lower bound, our baseline network with image alignment (D2) alone (\nth{4}~data row, Table ~\ref{tab:addlabelAb}) increased the average Dice to $85.0\%$ (1.7\% improvement compared to heavy augmentation). It shows that with reducing uncertain regions via entropy maps in the target domain, the target images have been brought closer to the source domain successfully. Then we add the output space alignment (D1) in the semantic prediction space, which slightly improved the average Dice (1.7\% improvement) and also achieved a good performance jump for the HD value  (\nth{5}~data row, Table~\ref{tab:addlabelAb}). Finally, by adding the point-could shape alignment network, the model was able to obtain an average Dice of $86.0\%$ (\nth{8}~data row, Table~\ref{tab:addlabelAb}), which, if compared with W/o UDA + augmentation, improved the segmentation performance by 3.3\%. The incremental increase in segmentation accuracy explains that feature adaptation and incorporating shape prior can be jointly conducted to achieve better domain adaptation.  Feature alignment in different compact spaces could inject effects from integral aspects to encourage domain invariance. The last row of the Table~\ref{tab:addlabelAb} shows the Dice and HD values, when we train our model in a multi-task fashion without any discriminators and adversarial learning. 

A very similar observation was achieved for the second benchmark dataset (MRI $\rightarrow$ CT). However, it is noticeable that the anatomy variation between MRI and CT modalities is quite significant, and it is a more challenging task compared to multi-sequence UDA. Table \ref{tab:ctmrtab} summarised the ablation study for MM-WHS dataset. In this experiment, the model with only entropy minimisation (D2) achieved an average volumetric Dice of $53.3\%$ and ASD volumetric value of $12.86mm$ (\nth{3}~data row, Table~\ref{tab:ctmrtab}), which is considerably lower than the model trained with output space alignment (D1) with an average Dice of $65.6\%$ and ASD of $5.97mm$ (\nth{2}~data row, Table~\ref{tab:ctmrtab}). Furthermore, when the model trained by employing both $\mathbf{D}_{1}+\mathbf{D}_{2}$ (\nth{4}~data row, Table~\ref{tab:ctmrtab}) discriminators outperformed the individual models.  It proves our initial hypothesis that enforcing entropy minimisation as an extra layer of adaptation can lead to a better model generalisation. Additionally, the model with point-cloud adaption (D3, \nth{5}~data row) without any other discriminators achieved comparable performance to $\mathbf{D}_{1}+\mathbf{D}_{2}$ model. By comparing these two results, we can conclude that shape adaptation alone has a significant role for UDA when there is a high variation between source and target domains.  Finally, the model with $\mathbf{D}_{1} + \mathbf{D}_{2} + \mathbf{D}_{3}$ outperformed the rest and achieved the best Dice and ASD scores across all cardiac structures. This model achieved an average volumetric Dice score of $72.5\%$ and an ASD score of $4.56 mm$, and improved the segmentation accuracy drastically, with a $22\%$ jump compared to W/o UDA method.

The results for the ablation study (MM-WHS dataset) highlights the contribution of adapting features in different spaces by our proposed model. There is an incremental improvement in terms of evaluation metrics after adding every discriminator. On the other hand, our proposed method performs only feature adaptation compared to other models such as SIFA (and CycleGAN) that applied both image and feature adaptation concurrently. Moreover, SIFA and CycleGAN methods translate the appearance of the source domain images to the target domain for UDA, while our model does not need this step and therefore has less trainable parameters. Image-based UDA methods highly depend on the quality of the image translation and require more data to train the whole network, which is always not accessible. Hence, methods such as the proposed one can be an alternative solution when there is a lack of sufficient training samples.

 \begin{table*}[ht]
   \centering
   \caption{Ablation study. Effectiveness of different adversarial domain adaptation components in our proposed framework for MS-CMRSeg dataset.}
      \resizebox{\textwidth}{!}{\begin{tabular}{ccccc|cccc|cccc}
     \hline
     \multicolumn{13}{c}{\textbf{Cardiac bSSFP-MRI $\rightarrow$ Cardiac LGE-MRI}} \\
     \hline
     Methods & \multicolumn{4}{c|}{Adversarial elements} & \multicolumn{4}{c|}{Volumetric Dice [\textit{mean~$\pm$~std}] $\uparrow$} & \multicolumn{4}{c}{Volumetric HD [mm] $\downarrow$} \\
    \cline{2-13}     \multicolumn{1}{c}{} & $\mathbf{D_{1}}$ & $\mathbf{D_{2}}$ & $\mathbf{D_{3}}$ & Aug & \multicolumn{1}{c}{Myo} & \multicolumn{1}{c}{LV} & \multicolumn{1}{c}{RV} & Average Dice & \multicolumn{1}{c}{Myo} & \multicolumn{1}{c}{LV} & \multicolumn{1}{c}{RV} & Average HD \\
     \hline \hline
     \multicolumn{1}{c}{} & $\times$ & $\times$ & $\times$ & $\times$ & \multicolumn{1}{c}{0.392$\pm$0.210} & \multicolumn{1}{c}{0.651$\pm$0.265} & \multicolumn{1}{c}{0.619$\pm$0.270} & 0.554$\pm$0.248 & \multicolumn{1}{c}{20.540$\pm$7.582} & \multicolumn{1}{c}{15.880$\pm$10.723} & \multicolumn{1}{c}{22.320$\pm$17.99} & 19.584$\pm$12.101 \\
     \multicolumn{1}{c}{W/o UDA} & $\times$ & $\times$ & $\times$ &$\checkmark$ & \multicolumn{1}{c}{0.577$\pm$0.161} & \multicolumn{1}{c}{0.797$\pm$0.141} & \multicolumn{1}{c}{0.755$\pm$0.126} & 0.710$\pm$0.143 & \multicolumn{1}{c}{20.950$\pm$10.289} & \multicolumn{1}{c}{19.279$\pm$8.858} & \multicolumn{1}{c}{15.946$\pm$7.855} & 18.728$\pm$9.001 \\
     \multicolumn{1}{c}{} & $\times$ & $\times$ & $\times$ &$\checkmark$ & \multicolumn{1}{c}{0.735$\pm$0.086} & \multicolumn{1}{c}{0.900$\pm$0.037} & \multicolumn{1}{c}{0.863$\pm$0.067} & 0.833$\pm$0.063 & \multicolumn{1}{c}{12.378$\pm$4.319} & \multicolumn{1}{c}{8.123$\pm$3.524} & \multicolumn{1}{c}{9.580$\pm$0.383} & 10.027$\pm$3.895 \\
     \hline
     \multicolumn{1}{c}{} &  $\times$ & $\checkmark$ & $\times$ &  $\checkmark$ & \multicolumn{1}{c}{0.778$\pm$0.061} & \multicolumn{1}{c}{0.906$\pm$0.034} & \multicolumn{1}{c}{0.867$\pm$0.063} & 0.850$\pm$0.053 & \multicolumn{1}{c}{9.727$\pm$2.686} & \multicolumn{1}{c}{7.375$\pm$3.311} & \multicolumn{1}{c}{9.952$\pm$3.459} & 9.018$\pm$3.152 \\
     \multicolumn{1}{c}{} &  $\checkmark$ &  $\checkmark$ & $\times$ &  $\checkmark$ & \multicolumn{1}{c}{0.776$\pm$0.07} & \multicolumn{1}{c}{0.904$\pm$0.04} & \multicolumn{1}{c}{0.869$\pm$0.067} & 0.85$\pm$0.059 & \multicolumn{1}{c}{11.483$\pm$4.538} & \multicolumn{1}{c}{8.005$\pm$2.977} & \multicolumn{1}{c}{9.509$\pm$3.946} & 9.666$\pm$3.82 \\
     \multicolumn{1}{c}{W/ UDA} & $\times$ & $\times$ &  $\checkmark$ &  $\checkmark$ & \multicolumn{1}{c}{0.769$\pm$0.046} & \multicolumn{1}{c}{0.906$\pm$0.035} & \multicolumn{1}{c}{0.866$\pm$0.067} & 0.847$\pm$0.049 & \multicolumn{1}{c}{11.258$\pm$3.215} & \multicolumn{1}{c}{7.888$\pm$3.588} & \multicolumn{1}{c}{9.780$\pm$5.033} & 9.642$\pm$3.945 \\
     \multicolumn{1}{c}{} &  $\times$ & $\checkmark$ &  $\checkmark$ &  $\checkmark$ & \multicolumn{1}{c}{0.789$\pm$0.039} & \multicolumn{1}{c}{0.907$\pm$0.03} & \multicolumn{1}{c}{0.877$\pm$0.053} & 0.858$\pm$0.041 & \multicolumn{1}{c}{10.01$\pm$3.457} & \multicolumn{1}{c}{7.682$\pm$2.700} & \multicolumn{1}{c}{8.987$\pm$3.989} & 8.893$\pm$3.382 \\
     \multicolumn{1}{c}{} &  $\checkmark$ &  $\checkmark$ &  $\checkmark$ &  $\checkmark$ & \multicolumn{1}{c}{\textbf{0.794$\pm$0.041}} & \multicolumn{1}{c}{\textbf{0.909$\pm$0.032}} & \multicolumn{1}{c}{\textbf{0.878$\pm$0.053}} & \textbf{0.860$\pm$0.042} & \multicolumn{1}{c}{\textbf{9.390$\pm$2.628}} & \multicolumn{1}{c}{\textbf{7.647$\pm$3.231}} & \multicolumn{1}{c}{\textbf{8.452$\pm$2.831}} & \textbf{8.496$\pm$2.897} \\
     \hline
     Multi-Task & $\times$ & $\times$ & $\times$ & $\checkmark$ & 0.78$\pm$0.049 & 0.908$\pm$0.028 & 0.872$\pm$0.053 & 0.853$\pm$0.043 & 10.140$\pm$3.136 & 7.710$\pm$2.908 & 10.007$\pm$3.649 & 9.286$\pm$3.231 \\
     \hline
     \end{tabular}}
   \label{tab:addlabelAb}
 \end{table*}%

\begin{table*}[ht]
  \centering
  \caption{Ablation study. Effectiveness of different adversarial domain adaptation components in our proposed framework for MM-WHS dataset (MRI $\rightarrow$ CT).}
    \begin{tabular}{lllll|ccccc|ccccc}
    \hline
    \multicolumn{15}{c}{\textbf{Cardiac MRI $\rightarrow$ Cardiac CT}} \\
    \hline
    Methods & \multicolumn{4}{c|}{Adversarial elements} & \multicolumn{5}{c|}{Volumetric Dice $(mean)$ $\uparrow$} & \multicolumn{5}{c}{Volumentric ASD $[mm]$ $\downarrow$} \\
    \cline{2-15}          &\textbf{ D1}    & \textbf{D2}    & \textbf{D3}    & Aug   & AA    & LA    & LV    & Myo   & Average & AA    & LA    & LV    & Myo   & Average \\
    \hline\hline
    W/o UDA &  $\times$     &   $\times$   &    $\times$   & $\times$    & 0.303 & 0.846 & 0.360 & 0.517 & 0.507 & 18.74 & 2.30  & 13.10 & 7.43  & 10.39 \\
    \hline

          &   $\checkmark$    &     $\times$   &    $\times$   & $\checkmark$    & 0.566 & 0.754 & 0.748 & 0.555 & 0.656 & 11.92 & 4.24  & 3.55  & 4.18  & 5.97 \\
          &   
          $\times$     &   $\checkmark$    &    $\times$     & $\checkmark$    & 0.812 & 0.765 & 0.329 & 0.225 & 0.533 & 3.68  & 3.63  & 15.41 & 28.71 & 12.86 \\
          &    $\checkmark$   &     $\checkmark$  &    $\times$   & $\checkmark$    & 0.794 & 0.782 & 0.583 & 0.571 & 0.683 & 4.42  & 3.37  & 8.64  & 5.44  & 5.47 \\
              W/ UDA
          &   $\times$     &     $\times$   &   $\checkmark$  & $\checkmark$    & 0.720 & 0.806 & 0.646 & 0.530 & 0.676 & 4.40  & 2.89  & 5.11  & 6.36  & 4.69 \\
          &    $\times$   &   $\checkmark$    &   $\checkmark$    & $\checkmark$    & 0.815 & 0.805 & 0.662 & 0.569 & 0.713 & 3.24  & 2.74  & 6.32  & 6.32  & 4.66 \\
          &    $\checkmark$   &    $\checkmark$    &    $\checkmark$    & $\checkmark$   & \textbf{0.830} & \textbf{0.813} & \textbf{0.672} & \textbf{0.584} & \textbf{0.725} & \textbf{2.90} & \textbf{2.66} & \textbf{6.32} & \textbf{6.38} & \textbf{4.56} \\
    \hline
        Multi-task &    $\times$   &   $\times$    &   $\times$    & $\times$    & 0.787 & 0.744 & 0.315 & 0.335 & 0.545 & 3.94  & 2.94  & 43.18 & 12.00 & 15.51 \\
    \hline
    \end{tabular}%
  \label{tab:ctmrtab}%
\end{table*}%

\subsubsection{Statistical Analysis} 
\begin{figure*}[ht]
    \centerline{\includegraphics[width=\textwidth]{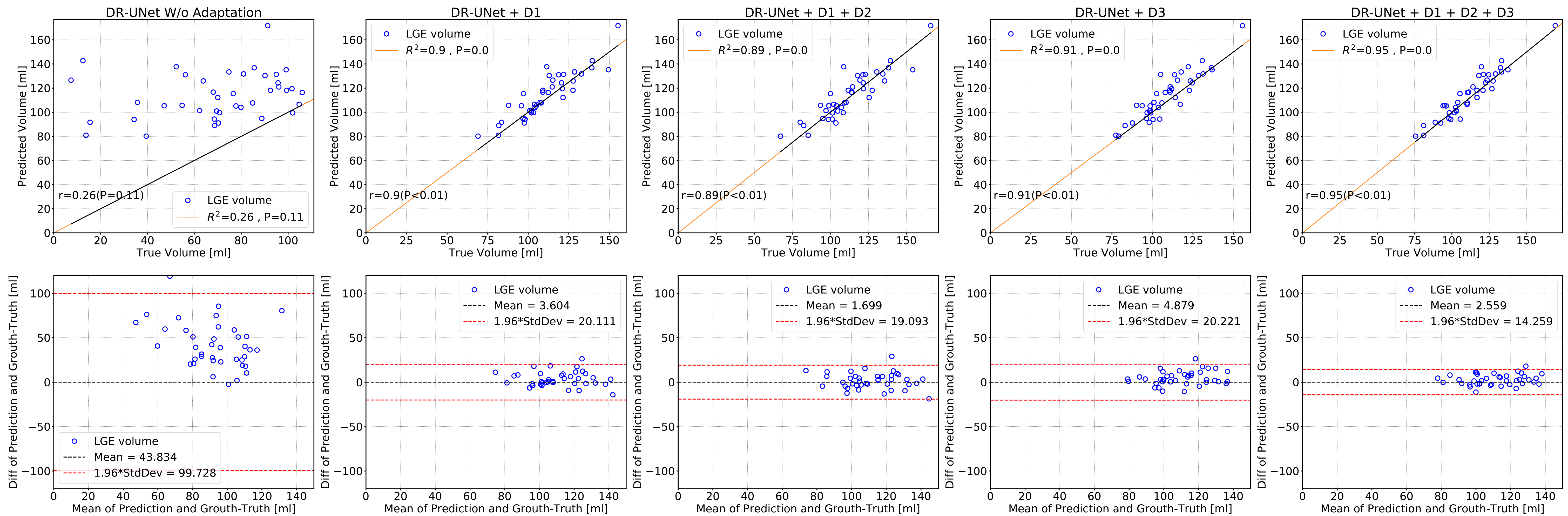}}
    \caption{Linear regression (top row) and Bland–Altman (bottom row) plots for the assessment of inter-observer variability for ventricular volume measurements by different methods. The volume measurements present an exemplary correlation $(R^{2} = 0.95)$ with a P-value $<0.01$ between our proposed UDA method (last column) and expert cardiologist annotation. Bland-Altman analysis plots show that the estimated ventricular volumes using our model is very close to the ground-truth (95\% confidence intervals). } \label{figBACP}
    \vspace{-0.5em}
\end{figure*}

To statistically evaluate the correlation and agreement between our proposed UDA and ground-truth segmentation generated by experts, linear regression and Bland-Altman plots are used. The linear regression plots are linear fits to the scatter plot of the true values vs. the predicted data for LV volume. The linear regression plot (\nth{1}~row,~Fig.~\ref{figBACP}) showed that Pearson's correlation is superior ($r^{2}=0.95$) for our proposed approach, \textit{DR-UNet + D1 + D2 + D3}, which includes entropy minimisation, point-cloud shape adaptation and output space alignment (P  $<0.01$ for all analyses, except W/o adaptation which achieved P-value of $0.11$). These results show that our method has substantial correlations with ground-truth segmentation by cardiologists. The Bland Altman plot shows limits of agreement (1.96 $*$ SD) at about 22 ml. These plots (\nth{2}~row,~Fig.~\ref{figBACP}) demonstrated that the 95\% limits of the agreement (14.25  mL) for our proposed method (\textit{DR-UNet + D1 + D2 + D3}) are much higher, which asserts that our method is closer to the actual annotations. The similar observation has been achieved for MM-WHS test data, but due to space limitations, we have not included the plots.

\section{Discussion}
For the diagnosis of cardiovascular diseases, several type of imaging modalities are used to measure the heart anatomy and morphology.  Deep learning techniques demonstrated significant performance in obtaining accurate and reliable segmentation of multi-modal cardiac images. Nevertheless, these networks demand labelled data for each modality because deep models are not able to generalise well across modalities due to differences in data distributions. Data annotation problem was addressed with multi-modal image-to-image translation~\cite{Zhu2017UnpairedCYCLEGAN, CYCADA, zhang2018translating} to generate synthesised images and shift the appearance of the target images to the source domain.  Recent studies investigated the application of cross-modality UDA to adapt deep features from the annotated source domain to the unlabelled target domain, achieving significant performances in computer vision~\cite{TsaiCVPR2018,vu2019advent} as well as in medical field~\cite{chen2019synergistic, CEFA2019, bermudez2018domain, Chen2019UnsupervisedMS}.  With a similar theme, our shape-aware and entropy minimisation method manifests that, without further annotation, the UDA segmentation method can significantly diminish the domain shift and could achieve similar performance as semi/fully supervised methods. 

In this paper, we present a novel interpretable UDA framework for multi-modal cardiac image segmentation. Our method integrates the entropy and point-clouds to leverage cross-modality priors from the source domain and to exploit the shape information embedded in the latent space of the segmentation network. Both adversarial learning components assist the model to learn invariant features for high entropy region and prior shape. Through multi-class segmentation tasks, we adequately demonstrated the overall effectiveness of our method without particular network architecture or hyperparameter tuning. It has been shown that a typical segmentation model like UNet or DR-UNet without domain adaptation has a high cross-modality performance degradation on the LGE-MRI and CT domains.  However, our method achieved significant improvement in the segmentation for the cross-modality tasks. 

\begin{figure}[t!]
    \centering
     \frame{\includegraphics[width=0.85\linewidth,height=0.30\linewidth]{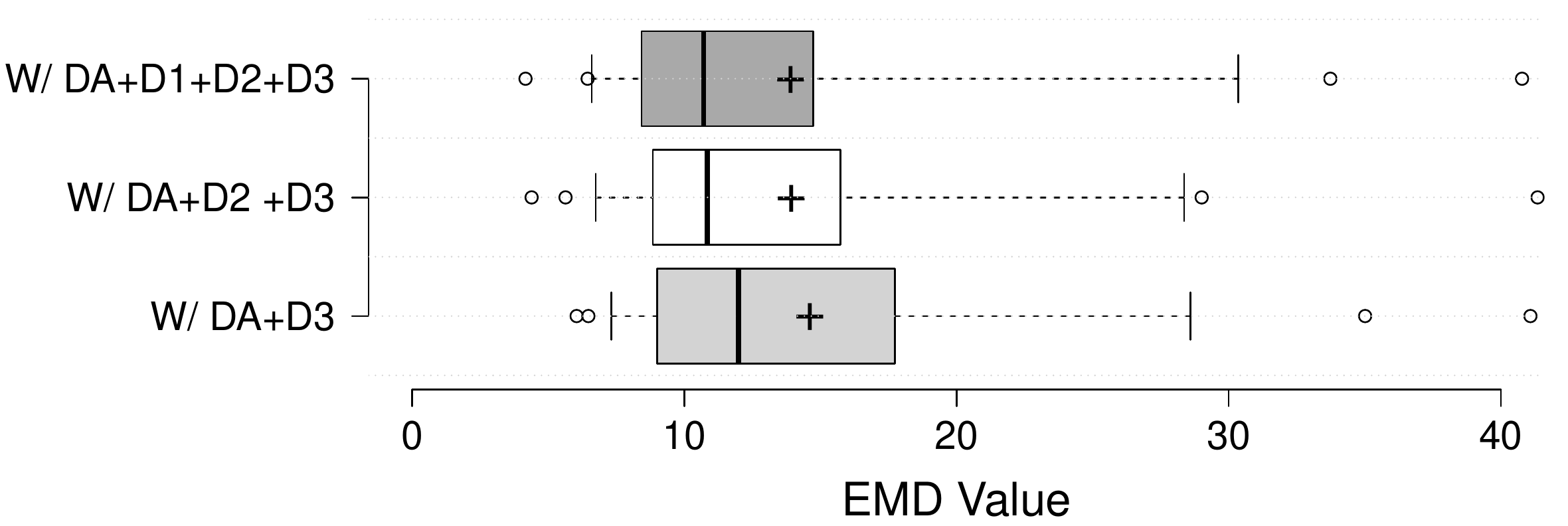}}
    \caption{Box-Plot of point-cloud EMD results in each ablation experimental setting for analysis of the proposed UDA framework.} \label{figPoint_emd}
        \vspace{-1.0em}
\end{figure}

\begin{figure*}[ht]
    \subfloat[]{\includegraphics[width=0.25\textwidth]{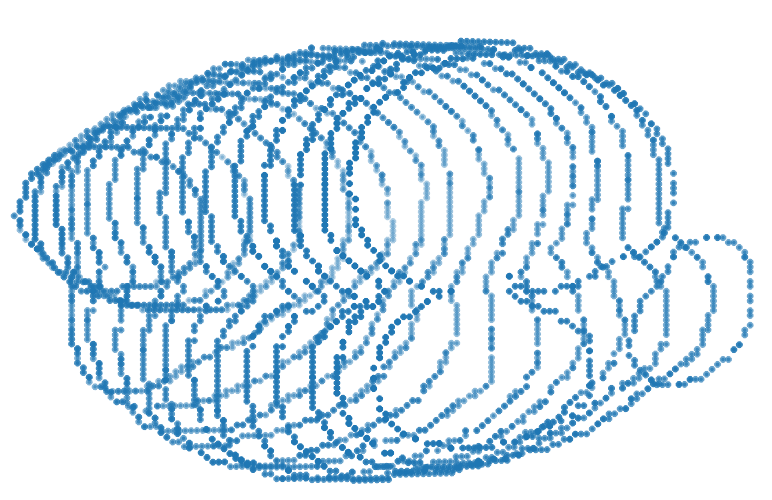}}
    \subfloat[]{\includegraphics[width=0.25\textwidth]{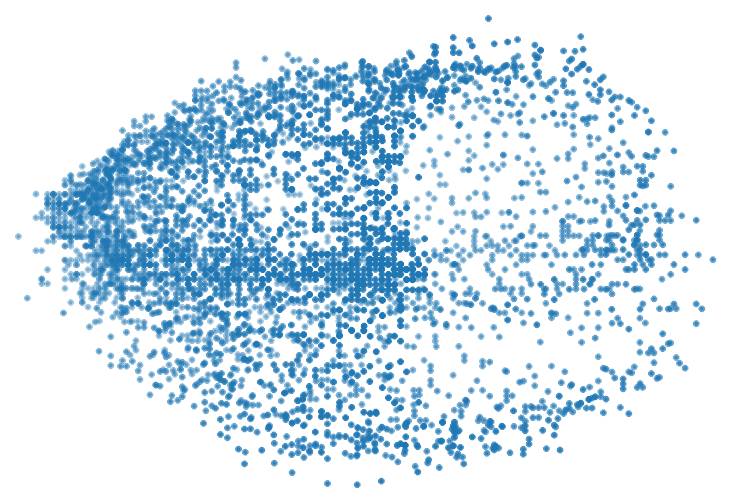}}
    \subfloat[]{\includegraphics[width=0.25\textwidth]{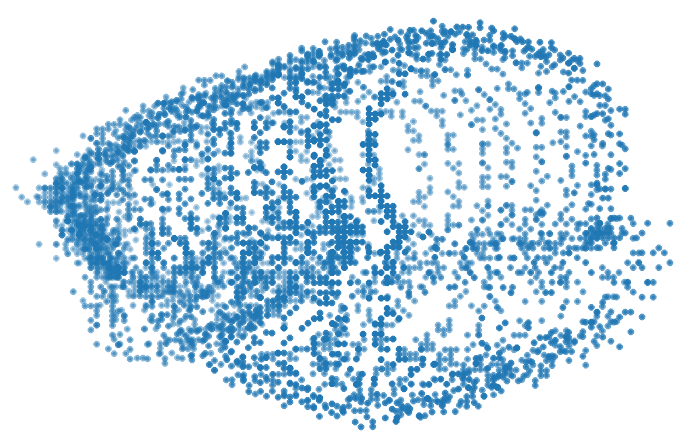}}
    \subfloat[]{\includegraphics[width=0.25\textwidth]{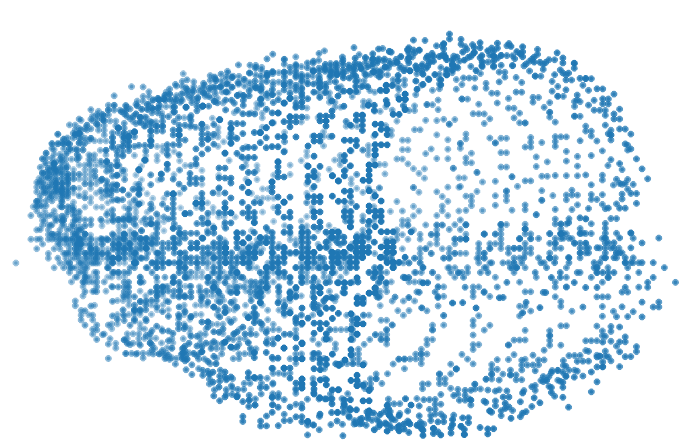}}
    \caption{3D visualisation of predicted point-clouds by our methods for LGE-MRI test data, after domain adaptation. (a) shows the ground-truth point-cloud, and (b-d) are the predictions for DR-UNet+D3, DR-UNet+D1+D2, and DR-UNet+D1+D2+D3, respectively.}
    \label{fig:animals}
\end{figure*}

The model (trained with an unsupervised learning strategy) that achieved the best Dice value of $87.3\%$~(Table~\ref{tab:addlabel}) on the MS-MWSeg benchmark is not trained end-to-end. The authors generate stylised LGE-MRI images using MUNIT \cite{huang2018multimodal}, thereby increasing the training data 10-fold. Then, they used a complex cascade UNet network architecture to refine their results. However, our method achieved a Dice value of $86.0\%$, using pure adversarial learning. The performance is not only similar to supervised and unsupervised models, but our model also outperformed all the methods by achieving the lowest HD value. The HD metric computes the smoothness of the object surface, and our point-clouds adaptation regressor could enforce the segmentation network in generating more smooth boundaries for ventricles. We further evaluated the accuracy of point-clouds regressor on the target domain (LGE-MRI) using EMD distance metric to see the impact of adversarial learning on three proposed methods~(Fig.~\ref{figPoint_emd}). \textit{DR-UNet + D1 + D2 + D3} achieved the lowest EMD distance in comparison. Furthermore, for qualitative evaluation, we visualised the 3D point-clouds by stacking the model prediction for a test LGE-MRI case~(Fig.~\ref{fig:animals}). A similar observation can be seen here as the generated point-clouds by \textit{DR-UNet + D1 + D2 + D3} has lower distance error in comparison to the ground-truth, and the overall shape of ventricles is well-preserved.

\begin{table*}[htbp]
  \centering
  \caption{ Slice-wise accuracies of the apical (Apex), mid-ventricular (Mid), and basal (Base) slices for different unsupervised methods.}
    \begin{tabular}{lcccc|cccc}
    \hline
    Methods & \multicolumn{4}{c|}{Slice-wise Dice [\textit{mean~$\pm$~std]} $\uparrow$} & \multicolumn{4}{c}{Slice-wise HD [mm] $\downarrow$} \\
\cline{2-9}          & Apex  & Mid   & Base  & Average & Apex  & Mid   & Base  & Average \\
    \hline\hline
    Baseline (W/o DA) & 0.479$\pm$0.260 & 0.216$\pm$0.214 & 0.191$\pm$0.232 & 0.295$\pm$0.235 & 7.94$\pm$16.67 & 32.23$\pm$16.19 & 26.41$\pm$24.17 & 22.19$\pm$19.01 \\
    \hline
    Chen et al.~\cite{Chen2019UnsupervisedMS} & \textbf{0.706$\pm$0.203} & \textbf{0.883$\pm$0.078} & \textbf{0.857$\pm$0.140} & \textbf{0.815$\pm$0.140} & 11.24$\pm$10.73 & \textbf{6.140$\pm$5.14} & \textbf{8.22$\pm$7.30} & \textbf{8.53$\pm$7.72 } \\
    Proposed method & 0.628$\pm$0.162 & 0.862$\pm$0.080 & 0.753$\pm$0.258 & 0.748$\pm$0.167 & \textbf{5.22$\pm$4.57} & 8.41$\pm$3.91 & 17.20$\pm$13.97 & 10.28$\pm$7.48 \\
    ADVENT~\cite{vu2019advent} & 0.623$\pm$0.237 & 0.859$\pm$0.086 & 0.746$\pm$0.205 & 0.743$\pm$0.176 & 5.49$\pm$3.68 & 9.10$\pm$5.02 & 20.47$\pm$15.27 & 11.69$\pm$7.99 \\
    Wang et al.~\cite{Wang22019} & 0.502$\pm$0.292 & 0.844$\pm$0.136 & 0.828$\pm$0.165 & 0.725$\pm$0.198 & 20.76$\pm$17.64 & 7.604$\pm$8.77  & 8.99$\pm$7.75 & 12.45$\pm$11.38 \\
    Ly et al.~\cite{Ly} & 0.612$\pm$0.230 & 0.795$\pm$0.142 & 0.724$\pm$0.221 & 0.710$\pm$0.198 &  14.45$\pm$15.52 & 12.74$\pm$13.57 & 16.99$\pm$15.93 & 14.72$\pm$15.00 \\
    \hline
    \end{tabular}%
  \label{tab:slicewise}%
\end{table*}%

Although the segmentation accuracy of our approach was high for the LV and RV blood pool, there is still room for improvement for the myocardium. In a few cases within the LGE-MRI dataset, the different myocardial wall and LV blood pool are obscure due to existing scars and edema in myocardium regions. A few such training cases makes the segmentation task more challenging.  Nevertheless, our point-cloud classifier shows an interesting ability where it is able to detect the borders between left and right ventricles (\nth{2}~row, \nth{4}~column, Fig.~\ref{figPoint_cloud}). Improving the point-cloud estimation accuracy from binary to multi-class regression may help in such cases to show some distinction between the epicardium and endocardium wall, which could lead to an improvement in the segmentation network accuracy, respectively. 

To have a more in-depth evaluation of the performance of our model, we computed the slice-wise segmentation accuracy from different positions of the ventricles, including the apical slices (Apex), mid-ventricular slices (Mid) and basal slices (Base) \cite{bernardTMI,zhuang2020cardiac}. Table \ref{tab:slicewise} demonstrates the slice-wise Dice and HD values that were computed, by averaging all the numbers from the Myo, LV, and RV cardiac structures. It can be observed that different methods achieved different performance regarding segmentation accuracy for the slices from different positions. The baseline (W/o) UDA showed the best performance on apex slices but achieved the lowest Dice score and highest HD for the middle and base slices.

Similar to MS-CMRSeg dataset, incremental improvement can also be observed on MM-WHS dataset by applying different discriminators for our proposed model. The boundaries between the cardiac structures in CT images are highly obscure compared to MR images making it hard to generalise the model trained on MR to CT images. Different from MS-MWSeg dataset, the difference in ablation study on cross-modality task is more significant. By comparing the baseline method W/o UDA and $\mathbf{D}_{2}$, we observe $\mathbf{D}_{2}$ generally performs better than W/o UDA, while $\mathbf{D}_{2}$ has higher variance. This poor performance can also be found in Fig.~\ref{fig:mrct_seg_etp}. The model with $\mathbf{D}_{2}$ has a completely wrong prediction on some of the substructures. By comparing the entropy rows, we find entropy is successfully reduced by applying $\mathbf{D}_{2}$. Although, a low entropy does not represent a better segmentation. Due to the large distribution discrepancy, prediction on the target domain should be poor. Minimising entropy of prediction with low certainty may encourage the model to generate wrong segmentations. However, the poor performance by applying entropy discriminator alone is addressed by utilising output-space and point-cloud discriminators, \textit{i.e.}, column $\mathbf{D}_{1} + \mathbf{D}_{2}$, $\mathbf{D}_{2} + \mathbf{D}_{3}$ and $\mathbf{D}_{1} + \mathbf{D}_{2} + \mathbf{D}_{3}$.

In volumetric image segmentation, incorporating temporal information (all slices put together) could be useful. Since our entire pipeline is based on 2D CNN models, the temporal information (present in the form of sequence of slices) is not integrated while training. Nonetheless, our proposed method is able to use the temporal information in the form of point-clouds as prior information for adversarial learning. Our UDA segmentation has several components, including the adversarial training part, and implementing such a network under the same configuration in 3D (using temporal information) requires a high amount of memory and longer training time. Nevertheless, we believe that incorporating shape information using point-clouds in the 3D domain could lead to better performance improvement since the overall shape of cardiac anatomy will be taken into consideration in a temporal manner.  

\section{Conclusion}
In this paper, we proposed a novel entropy and shape-aware UDA method for multi-modal cardiac MR image segmentation. We approached the domain adaptation problem via adversarial learning in different spaces.  We showed that introducing additional shape information using point-clouds along with entropy minimisation brings further complementary effects to bridge the performance gap between source and target domains. We construct a novel segmentation network such that the point-cloud information is embedded into a dedicated deep architecture using an auxiliary point-cloud regression task. A novel point-cloud discriminator based on PointNet is proposed to distinguish whether the point-clouds are from source or target domain.  Experimental results on two benchmark cardiac image datasets highlighted that the proposed end-to-end method outperforms many baseline models and SOTA algorithms. For future work, we aim to extend our UDA segmentation method from 2D to 3D to process volumetric data for adaptation,  which is clinically more relevant. To prove the generalisability and robustness of our method, we plan to test it on larger clinical cohort studies and employ it on other domains beyond cardiac segmentation task, such as multi-modal brain studies for lesion segmentation. We also aim to find a better way to choose the hyperparameters $\lambda_1$, $\lambda_2$ and $\lambda_3$ for the total loss function and learning rates, especially on how to efficiently control the effect of the gradient with respect to different discriminators for better model convergence. 

\bibliographystyle{IEEEtran}
\bibliography{biblo}
\end{document}